\documentclass[10pt,letterpaper]{article}
\ifx\XeTeXversion\undefined\pdfoutput=1\pdfminorversion=7\fi
\usepackage[T1]{fontenc}
\usepackage[utf8]{inputenc}
\usepackage{lmodern}
\usepackage[letterpaper,left=0.90in,right=0.90in,top=0.85in,bottom=0.85in]{geometry}
\usepackage{amsmath,amssymb,graphicx,booktabs,array}
\usepackage[super,sort&compress,comma]{natbib}
\usepackage{microtype}
\usepackage{xcolor}
\definecolor{ink}{HTML}{000000}
\definecolor{rulegray}{HTML}{CCCCCC}
\definecolor{teal}{HTML}{0A6264}
\usepackage[colorlinks=true,linkcolor=teal,citecolor=teal,urlcolor=teal,pdftitle={EnzDecomp: Disentangling background reactivity from active-site selectivity in enzyme screens},pdfauthor={Maohua Xu, Xukun Qin}]{hyperref}
\providecommand{\doi}[1]{\href{https://doi.org/#1}{\nolinkurl{#1}}}
\makeatletter
\long\def\@makecaption#1#2{%
  \vskip\abovecaptionskip
  \sbox\@tempboxa{{\small\textbf{#1\ |\ }#2}}%
  \ifdim \wd\@tempboxa >\hsize
    {\small\textbf{#1\ |\ }#2\par}
  \else
    \global \@minipagefalse
    \hb@xt@\hsize{\hfil\box\@tempboxa\hfil}%
  \fi
  \vskip\belowcaptionskip}
\renewcommand{\section}{\@startsection{section}{1}{\z@}%
  {-3.0ex \@plus -0.8ex \@minus -0.2ex}%
  {1.3ex \@plus 0.2ex}%
  {\normalfont\Large\bfseries\raggedright\hyphenpenalty=10000\exhyphenpenalty=10000\color{ink}}}
\renewcommand{\subsection}{\@startsection{subsection}{2}{\z@}%
  {-2.2ex \@plus -0.6ex \@minus -0.2ex}%
  {0.9ex \@plus 0.2ex}%
  {\normalfont\large\bfseries\raggedright\hyphenpenalty=10000\exhyphenpenalty=10000\color{ink}}}
\renewcommand{\maketitle}{%
 {\raggedright\hyphenpenalty=10000\exhyphenpenalty=10000
  {\LARGE\bfseries\color{black}\@title\par}%
  \ifx\@author\@empty\else\vspace{0.65em}{\normalsize\color{black}\hypersetup{urlcolor=black}\@author\par}\fi
 }\vspace{0.55em}}
\makeatother

\title{EnzDecomp: Disentangling background reactivity from active-site selectivity in enzyme screens}
\author{%
{\large\bfseries Maohua Xu\textsuperscript{1} \& Xukun Qin\textsuperscript{2,3,\ensuremath{\boldsymbol{\ast}}}}\\[0.45em]
{\small
\textsuperscript{1}\,Department of Biomedical Engineering, The Pennsylvania State University, University Park, PA, USA\\[0.15em]
\textsuperscript{2}\,Ocean University of China, Qingdao, China\\[0.15em]
\textsuperscript{3}\,Shanghai Artificial Intelligence Laboratory, Shanghai, China\\[0.28em]
\textsuperscript{\ensuremath{\ast}}\,Corresponding author: \href{mailto:xkqin@stu.ouc.edu.cn}{\nolinkurl{xkqin@stu.ouc.edu.cn}} (X.Q.);\enspace e-mail: \href{mailto:mzx5066@psu.edu}{\nolinkurl{mzx5066@psu.edu}} (M.X.)}%
}
\date{}

\begin{document}
\maketitle
\vspace{0.15em}
{\color{rulegray}\hrule height 0.6pt}
\vspace{0.55em}
\noindent{\large\bfseries\color{ink}Abstract}\par\vspace{0.35em}
{\noindent
Ethyl hexanoate and hexyl acetate differ in which side of the ester carries the longer chain. Of 146 esterases, 28 cleaved ethyl hexanoate and 29 cleaved hexyl acetate. Both look similar if any catalyst suffices. Selective linkers instead rely on the ten enzymes specific to ethyl hexanoate and the eleven specific to hexyl acetate. We asked if both sorts of entries could be predicted for enzymes and substrates held out of training. The same issue arises with epimers and enantiomers, where a reversal of enzymatic preference may accompany a change in configuration without any change in connectivity. For each of six enzyme families we had a matrix of compounds against related enzymes, with every entry from the same assay, including inactive pairs.\par
An additive fit to these matrices already ranked substrates nearly as well as the best external model (within-enzyme average precision (AP) 0.468 versus 0.477 over 176 splits). Yet four SMILES-based predictors tied on 70.5--100\% of unseen epimer or enantiomer cases. EnzDecomp fits substrate reactivity and then enzyme breadth, freezing each in turn. No docking is needed. We fit the residual using a fixed 3D pocket per enzyme, local 2D chiral bond patterns and quantum-chemical descriptors at candidate reaction sites, then center it on the training matrix. On 4,355 reversals this yielded the best native balanced accuracy (61.6\%, rising to 80.1\% on esterases). Accuracy on 780 unseen isomer cases was 66.3\% (epimers 58.3\%, enantiomers 61.6\%, no ties). The full score led within-enzyme ranking on 152 non-random and 74 double-holdout splits (0.475 and 0.436 AP) and whole-library ranking on the HAD panel with $<$40\%-identity enzyme clusters (0.302 pooled AP). Downstream training took 10.4 minutes per esterase split, 25--44 times faster than the docking and multimodal Transformer baselines.\par}
\vspace{0.65em}
{\color{rulegray}\hrule height 0.6pt}

\section*{Introduction}
Suppose two ester bonds occur in the linkers of a responsive material. It matters whether the same enzyme cleaves both. Related choices of enzyme and substrate arise in phosphate and ester prodrugs, in glycosyltransferase-mediated remodeling of cell-surface glycans, and in the synthesis of chiral intermediates\citep{huang2015,martinez2018,yang2018}. Activity databases are of limited help with this question if two reactions were reported under different conditions, or if an enzyme--substrate combination recorded as negative has never been assayed\citep{esp,prosmith}. We used family screens in which tens to hundreds of homologs encountered the same substrates under uniform conditions. These experiments give measured positives and negatives for esterases\citep{martinez2018}, haloacid dehalogenase (HAD) phosphatases\citep{huang2015}, OleA thiolases\citep{robinson2020}, glycosyltransferases\citep{yang2018}, nitrilases\citep{black2015} and DUF849 $\beta$-ketoacid cleavage enzymes\citep{bastard2014}. On unfamiliar enzymes and chemical structures\citep{goldman2022}, screening asks which reactions to prioritize for one enzyme or across a panel. Selectivity asks whether an enzyme prefers $s$ over its isomer or enantiomer $t$ and whether a second enzyme prefers $t$.

Rankings within a family can shift when new enzyme homologs or chemical scaffolds are tested\citep{goldman2022,datasplit2026,mahood2026}. In a family screen, enzyme breadth, shared substrate reactivity and pocket interaction\citep{lengerich2020} can all three contribute to a hit. Differences in enzyme breadth (including expression differences) alter hit counts before specific pocket contacts enter. Activated esters or aryl phosphates with labile leaving groups raise hits across the family. When two substrates are compared on one enzyme, enzyme breadth cancels out, but the difference in shared substrate reactivity remains and favors the same substrate for every enzyme. Take the esters ethyl hexanoate and hexyl acetate. Both have the formula $\mathrm{C_8H_{16}O_2}$. Among 146 esterases tested by Mart{\'\i}nez-Mart{\'\i}nez et al.\citep{martinez2018}, 28 cleaved ethyl hexanoate and 29 cleaved hexyl acetate. Ten of the enzymes cleaved the former ester alone and 11 the latter alone. Thus even for two esters with nearly equal hit rates, the more active substrate depends on which enzyme is assayed. HAD phosphatases likewise show opposite preferences for epimers and for pairs of enantiomers\citep{huang2015}.

For a given enzyme and substrate, a protein--molecule encoder\citep{esp,enzrank,fusionesp,omniesi,puepp,rammesi}, a multimodal language model such as ProSmith\citep{prosmith}, and a graph network for a docked complex such as EZSpecificity\citep{ezspecificity} each return an activity score. How much of a good screening score actually requires a match between the two molecules? We scored each reaction with only enzyme breadth $A(e)$ and shared substrate reactivity $C(s)$. With \textbf{no enzyme--substrate interaction}, this additive score reached \textbf{0.468 within-enzyme average precision (AP)} on \textbf{176 splits}, about \textbf{0.8 percentage points} below the best external model (\textbf{0.477}), yet gave only \textbf{50.0\%} balanced accuracy on reversals. None of the \textbf{43/43 enantiomer pairs} could be told apart by their 1D SMILES embeddings. Compressing 2D descriptors by principal components lost nearly \textbf{70\%} of the separation among stereoisomers, and four predictors gave equal scores in \textbf{70.5--100\%} of enzyme-level comparisons on unseen epimers and enantiomers. A 3D docked representation can distinguish handedness, but encoding a screen in EZSpecificity takes $N_E\times N_S$ complexes. Even downstream training alone took \textbf{25$\times$} as long, and its output still contains shared activity.

\begin{figure}[!t]
\centering
\includegraphics[width=\linewidth]{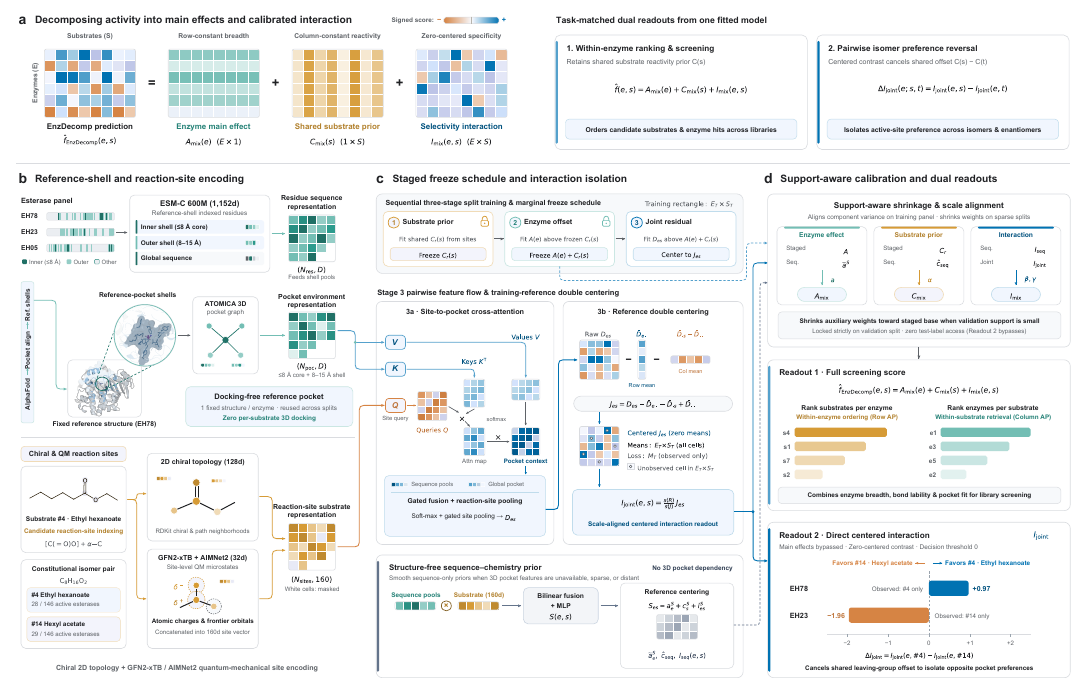}
\caption{\textbf{How one family screen supplies both predictions.} \textbf{a}, In the $E\times S$ matrix, $A_{\rm mix}(e)$ varies by enzyme and $C_{\rm mix}(s)$ by substrate. The third term $I_{\rm mix}(e,s)$ is the centered interaction. The complete score ranks reactions for a library screen. The direct interaction readout compares two substrates. \textbf{b}, Reference pocket of esterase EH78, rendered in PyMOL\citep{pymol}. For each enzyme we encode one structure with ATOMICA and the surrounding residues with ESM-C 600M (1,152 features per residue). At candidate reactive groups in each protonation microstate, we use 128 local RDKit channels for topology and chirality\citep{rdkit} and 32 GFN2-xTB/AIMNet2 channels for electronic structure. \textbf{c}, After fitting and freezing $C_r(s)$ in Stage~1 and $A(e)$ in Stage~2, we fit the remaining variation using attention between sites and the pocket with soft-maximum and gated pooling (\textbf{3a}). Double centering on the training rectangle $E_T\times S_T$ (\textbf{3b}) gives $I_{\rm joint}$. A sequence--chemistry prior covers sparse, absent or distant pocket structures. \textbf{d}, Variance alignment and support shrinkage on the validation split recombine terms for Readout~1 ($\hat f_{\rm EnzDecomp}$). Readout~2 ($I_{\rm joint}$) takes the interaction directly.}
\label{fig:architecture}
\end{figure}

EnzDecomp takes one reference region from each enzyme, encoding the residues around it with ESM-C\citep{esmc} and the 3D pocket with ATOMICA\citep{atomica} (Fig.~\ref{fig:architecture}). For each substrate we enumerate candidate reactive groups and protonation microstates, using 128 channels of local 2D chiral topology\citep{rdkit} and 32 GFN2-xTB/AIMNet2 quantum channels\citep{gfn2xtb,aimnet2}. Stages~1 and~2 fit and successively freeze substrate reactivity and enzyme breadth. Stage~3 predicts the residual through site-to-pocket attention and double-centers it on the training matrix. A sequence--chemistry prior can be used when pocket structures are unavailable or come from distant homologs. Readout~1 ($\hat f_{\rm EnzDecomp}$) combines the components using validation data for library screening. Readout~2 ($I_{\rm joint}$) takes the Stage~3 interaction directly. Among nine models on 176 splits, the direct Stage~3 $I_{\rm joint}$ readout calls 19 of the 21 selective esterases correctly. Balanced accuracy was 80.1\% on esterase substrate pairs, 58.3\% on unseen epimers and 61.6\% on unseen enantiomers, with 0.0\% ties. We also double-centered each external score matrix on its training reference panel. Cross-panel balanced accuracy increased for all 8/8 external models (Figs.~\ref{fig:selectivity} and~\ref{fig:isomers}). The full score led within-enzyme ranking on 152 non-random and 74 double-holdout splits (0.475 and 0.436 AP) and whole-library ranking on the HAD panel with $<$40\%-identity enzyme clusters (0.302 pooled AP, Fig.~\ref{fig:had}). Training the identical network without freezing marginal terms gave 59.6\% versus 80.1\% on esterase substrate pairs. The relative spread of fitted enzyme and substrate effects varied by family, and quantum descriptors helped on new functional groups. Downstream training was 25--44 times faster than EZSpecificity and ProSmith (Fig.~\ref{fig:mechanism}).

\section*{Results}
\subsection*{Additive background reactivity drives black-box screening rankings}
A chemist looking for substrates of a single enzyme can rank the whole assayed library. One who wants two enzymes with opposite preferences must compare the same substrate pair across enzymes. The family screens allow both tests on the same held-out cells: they record active and inactive reactions for homologs tested against a common substrate library under common conditions. We compared EnzDecomp with eight published predictors, retraining each on the same \textbf{176 train--test splits}. The eight panels cover six enzyme families (esterases, HAD phosphatases, OleA thiolases, glycosyltransferases, nitrilases and DUF849 $\beta$-ketoacid cleavage enzymes; Table~\ref{tab:panels} and Methods). For screening we used \emph{within-enzyme substrate ranking}, measured by row-wise average precision (Row AP). For \emph{preference reversal balanced accuracy} we found \textbf{4,355 high-support substrate-pair cases} with opposite preferences among test enzymes, comprising 187,007 discordant enzyme--pair comparisons (Tables~\ref{tab:overall} and~\ref{tab:reversal}).

The four highest-ranking external predictors differed by just \textbf{0.8 percentage points} in Row AP (\textbf{0.469--0.477}, with AUROC of 0.780--0.794; Fig.~\ref{fig:mechanism}a). They use quite different inputs and architectures: ProSmith is a multimodal language model\citep{prosmith}, OmniESI uses progressive conditional fusion\citep{omniesi}, RAMMESI combines retrieval with feature fusion\citep{rammesi}, and EZSpecificity builds a graph of each docked complex\citep{ezspecificity}. Their balanced accuracies on preference reversals ranged from \textbf{56.8 to 57.8\%}. FusionESP reached 59.4\% on reversals with a lower Row AP of 0.427\citep{fusionesp}. We tried scoring a reaction using just the breadth of the enzyme $A(e)$ and the shared reactivity of the substrate $C(s)$. On reversals $A(e)+C(s)$ gives exactly \textbf{50.0\%} balanced accuracy. As a screening model, however, it gave \textbf{0.468 Row AP} and \textbf{0.773 AUROC} (Fig.~\ref{fig:mechanism}a and Table~\ref{tab:internal}). Its Row AP exceeded \textbf{four of the eight} external scores, nearly matched RAMMESI (0.469) and fell about \textbf{0.8 percentage points} below the highest external score (OmniESI, 0.477). With one score per pair, enzyme breadth and substrate reactivity dominate the fit over active-site differences. We checked their role by centering the eight external score matrices and comparing staged with unstaged training on the same network.

We next let the additive fit make an enzyme-specific correction for each substrate. The simplest correction was just a scalar times the centered Stage~3 residual, giving $A+C_r+\gamma J$. With no sequence-only prior and no post-hoc calibration, this three-stage model gave \textbf{0.478 Row AP} across the 176 splits, higher than each external score (Table~\ref{tab:internal}). For library screening, alignment on the validation data according to support brought the full score (Readout~1, $\hat f_{\rm EnzDecomp}$) to \textbf{0.483 Row AP}. For preferential activity, the quantity of interest was already available from the same fit: the pocket term $I_{\rm joint}$ (Readout~2). Used directly, it gave \textbf{61.6\%} balanced accuracy over the 4,355 reversal cases (Fig.~\ref{fig:architecture}c,d and Fig.~\ref{fig:mechanism}a). The full score for screening and the pocket term for reversals also led on these two metrics when single runs were scored separately and the scores averaged (Table~\ref{tab:fiverun}).

\subsection*{Removing shared substrate bias resolves preference reversals}
\begin{figure}[p]
\centering
\includegraphics[width=\linewidth,height=0.79\textheight,keepaspectratio]{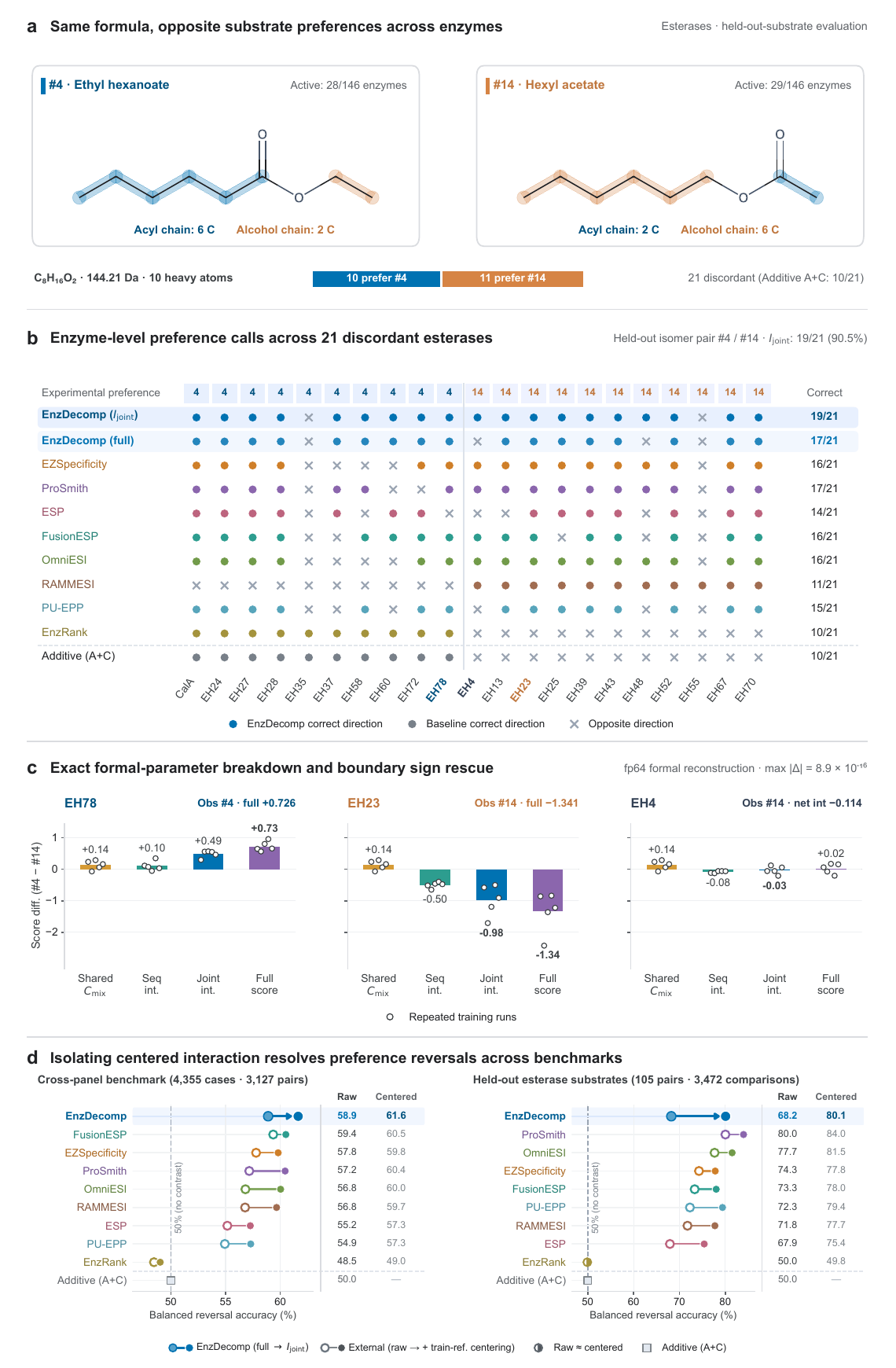}
\caption{\textbf{Resolving enzyme-specific preference reversals by removing shared substrate bias.} \textbf{a}, Ethyl hexanoate (\#4) and hexyl acetate (\#14) are held-out constitutional isomers ($\mathrm{C_8H_{16}O_2}$) with similar hit rates across $n=146$ esterases. Among the $n=21$ enzymes active on just a single isomer, 10 preferred \#4 and 11 preferred \#14. \textbf{b}, Predicted preference direction for all 21 selective esterases under substrate holdout. Filled circles mark correct calls, crosses mark opposite calls, and exact ties get zero credit. \textbf{c}, Exact 64-bit floating-point score breakdown over $n=5$ runs ($\max|\Delta|=8.9\times 10^{-16}$), where open circles show individual runs for strongly selective esterases EH78 and EH23 and boundary esterase EH4. Without the shared substrate offset $C_{\rm mix}=+0.136$, EH4 keeps its negative pocket preference for \#14 ($-0.114$ rather than $+0.022$ in the full score). \textbf{d}, Equal-case mean balanced reversal accuracy across all $n=4,355$ benchmark reversal cases (3,127 unique pairs; left) and $n=105$ held-out esterase pairs (3,472 comparisons; right), comparing EnzDecomp's full score and $I_{\rm joint}$ readout (blue circles) against external raw scores (open circles) and post-hoc training-reference centered scores (filled circles).}
\label{fig:selectivity}
\end{figure}

Ethyl hexanoate (\#4) and hexyl acetate (\#14) were both held out of training (Fig.~\ref{fig:selectivity}a). These two constitutional isomers share the formula $\mathrm{C_8H_{16}O_2}$. Ethyl hexanoate and hexyl acetate differ in which side of the ester carries the longer six-carbon chain. In the metagenomic esterase screen\citep{martinez2018}, 28 of 146 enzymes cleaved ethyl hexanoate and 29 cleaved hexyl acetate. Only 21 of the 146 esterases were active on one ester: ten cleaved ethyl hexanoate alone and eleven cleaved hexyl acetate alone. Even for esters with nearly equal family-wide hit rates, which isomer is hydrolyzed depends on which enzyme is assayed. Their opposite preferences reflect differences in pocket geometry near the catalytic triad. With $A+C$, the prediction was the same for all 21 enzymes (10/21 correct). External models gave at most \textbf{17 of 21} correct calls (Fig.~\ref{fig:selectivity}b and Table~\ref{tab:reversal}).

EH4 and EH48 both prefer hexyl acetate. In the 64-bit floating-point reconstruction, the sequence and 3D pocket terms favor that ester by $-0.114$ and $-0.129$, respectively (Fig.~\ref{fig:selectivity}c). The full score gets both preferences wrong ($+0.022$ and $+0.006$): each enzyme has the same $C_{\rm mix}=+0.136$ offset in favor of ethyl hexanoate. The offset is too small to change the calls for EH78 and EH23. Their pocket preferences for ethyl hexanoate ($+0.59$, EH78) and hexyl acetate ($-1.48$, EH23) retain their signs in the full scores ($+0.73$ and $-1.34$). The direct Stage~3 $I_{\rm joint}$ readout omits $C_{\rm mix}$, restores the negative signs for EH4 and EH48 and calls \textbf{19 of the 21} enzymes correctly (\textbf{90.5\%}; Fig.~\ref{fig:selectivity}b,c). Across 4,355 reversal cases the change from full score to interaction raised balanced accuracy from 58.9\% to \textbf{61.6\%}. On 105 held-out esterase substrate pairs the increase was from 68.2\% to \textbf{80.1\%}. These were the \textbf{highest native scores} on both benchmarks (Fig.~\ref{fig:selectivity}d and Table~\ref{tab:reversal}). We also double-centered each external score matrix on its training reference panel. Cross-panel balanced accuracy rose for \textbf{all 8/8 external models}, consistent with a shared reactivity bias. Centered FusionESP and ProSmith reached 60.5\% and 60.4\% overall, compared with \textbf{61.6\%} for native Stage~3 $I_{\rm joint}$, and centered ProSmith reached 84.0\% on the esterase screen (Fig.~\ref{fig:selectivity}d, filled circles).

\subsection*{Chiral reaction-site encoding resolves epimers and enantiomers}
\begin{figure}[p]
\centering
\includegraphics[width=\linewidth,height=0.77\textheight,keepaspectratio]{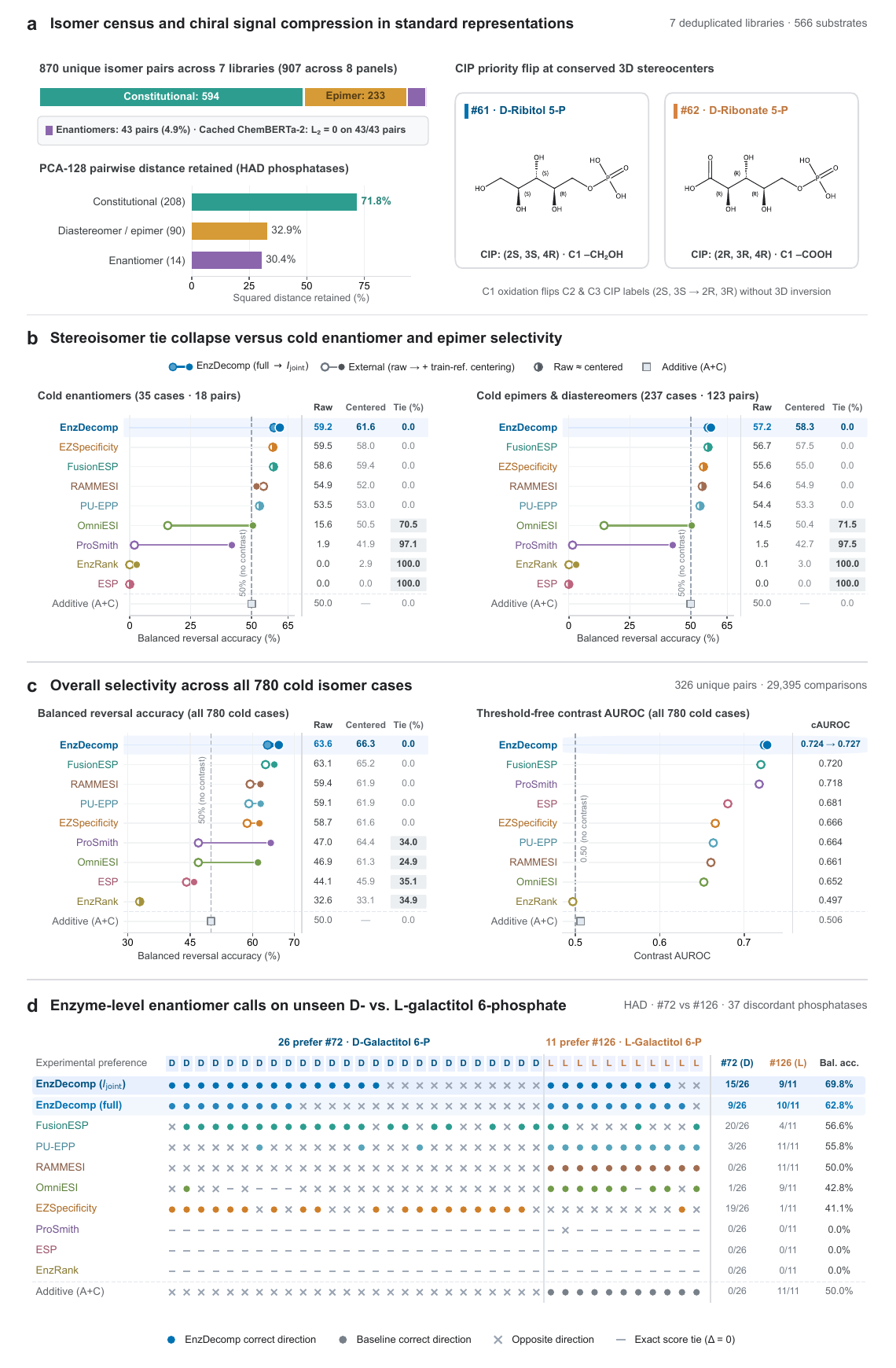}
\caption{\textbf{Selectivity across constitutional isomers, epimers and mirror-image enantiomers.} \textbf{a}, The seven deduplicated substrate libraries contain $n=870$ unique isomer pairs (907 across eight panels, top left). At bottom left, the SMILES embeddings have $L_2=0$ for 43/43 enantiomer pairs. The same panel shows how much stereochemical information was retained in the HAD phosphatase descriptors after PCA-128 compression. At right, aligned D-ribitol 5-phosphate (\#61) and D-ribonate 5-phosphate (\#62) retain the same 3D stereocenter geometry. Oxidation at C1 changes substituent priorities and hence the formal $(R/S)$ labels at C2 and C3. \textbf{b}, Balanced reversal accuracy and exact tie rates (\%) on $n=35$ unseen enantiomer cases (18 unique pairs, left) and $n=237$ unseen diastereomer or epimer cases (123 unique pairs, right). Blue circles show EnzDecomp, open circles raw external scores, and filled circles post-hoc centered scores. \textbf{c}, Overall balanced accuracy (left) and threshold-free contrast AUROC (right) across all $n=780$ unseen isomer cases (326 unique pairs; 29,395 discordant comparisons). \textbf{d}, Enantiomer calls across $n=37$ selective HAD phosphatases, 26 D-selective and 11 L-selective, on unseen D- versus L-galactitol 6-phosphate (\#72 versus \#126). Dashes mark exact score ties with $|\Delta f|\leq 10^{-12}$.}
\label{fig:isomers}
\end{figure}

A bond graph alone cannot distinguish stereoisomers with the same connectivity and different tetrahedral configurations. Across seven distinct substrate libraries we found 566 compounds and \textbf{870 unique isomer pairs} with matching molecular formulas. Constitutional isomers accounted for 594 pairs, diastereomers (including epimers) for 233 and mirror-image enantiomers for 43 (Fig.~\ref{fig:isomers}a). None of the 43 enantiomer pairs could be distinguished by distance between their ChemBERTa-2 SMILES embeddings. For every pair, $L_2=0$ (Fig.~\ref{fig:isomers}a). In the curated $218\times131$ HAD phosphatase panel, reducing the 2D descriptors to 128 principal components retained 71.8\% of the squared distance between constitutional isomers, but only 32.9\% between epimers and 30.4\% between enantiomers. The Cahn--Ingold--Prelog labels have a different limitation. Consider D-ribitol 5-phosphate (\#61, $2S,3S,4R$) and its C1 oxidation product, D-ribonate 5-phosphate (\#62, $2R,3R,4R$; Fig.~\ref{fig:isomers}a, right). Changing the terminal alcohol to a carboxyl group changes the substituent priorities at C2 and C3. Both labels change from $S$ to $R$, although the spatial arrangement at each stereocenter is conserved.

Substrate holdout produced \textbf{780 reversal cases} involving isomers (326 unique pairs and 29,395 discordant comparisons; Fig.~\ref{fig:isomers}b,c and Table~\ref{tab:isomers}). With chiral reaction-site features and cross-attention to the 3D pocket, $I_{\rm joint}$ produced \textbf{0.0\% ties}. It ranked first among native and post-hoc centered predictors on both balanced accuracy and threshold-free contrast AUROC: \textbf{66.3\%} and \textbf{0.727} overall (Fig.~\ref{fig:isomers}c), \textbf{58.3\%} and \textbf{0.629} for the 237 diastereomer/epimer cases, and \textbf{61.6\%} and \textbf{0.641} for the 35 mirror-image enantiomer cases (Fig.~\ref{fig:isomers}b). In the 508 constitutional-isomer cases the molecules had different 2D bond graphs. ProSmith attained 71.3\% raw and 76.0\% centered balanced accuracy on these cases, compared with 70.3\% for $I_{\rm joint}$. When the bond graphs were identical, ESP\citep{esp}, EnzRank\citep{enzrank}, ProSmith and OmniESI tied on \textbf{70.5--100\%} of unseen epimer and enantiomer comparisons. Their raw balanced accuracies ranged from \textbf{0 to 15.6\%} (Fig.~\ref{fig:isomers}b). On unseen mirror-image polyol phosphates D-galactitol 6-phosphate (\#72) versus L-galactitol 6-phosphate (\#126) across 37 selective HAD phosphatases (Fig.~\ref{fig:isomers}d), ESP and EnzRank output exact ties on all 37 enzymes and ProSmith on 36 of 37 (\textbf{0.0\%} accuracy); $A+C$ and RAMMESI predicted one direction for all 37 enzymes (50.0\%); and 3D-docked EZSpecificity captured 19/26 D-selective enzymes but only 1/11 L-selective enzymes (41.1\%), whereas $I_{\rm joint}$ resolved bidirectional stereoselectivity across both D- and L-selective cap-domain architectures (15/26 and 9/11; \textbf{69.8\% balanced accuracy}).

\subsection*{Cluster-isolated extrapolation across homologs and scaffolds}
For screening we used the full score (Readout~1, $\hat f_{\rm EnzDecomp}$), which combines calibrated enzyme breadth, substrate reactivity and pocket interaction. For pairwise selectivity we used Readout~2 ($I_{\rm joint}$). Across all 176 splits, the full score had the highest mean Row AP of the nine models (\textbf{0.483}; Table~\ref{tab:overall}). In the 24 random-pair splits, close homologs and chemical analogs of test pairs appeared in training. On those splits, top fusion networks scored 0.538--0.543 and EnzDecomp scored 0.535. Across the \textbf{152 non-random splits}, EnzDecomp led at \textbf{0.475 Row AP}, ahead of OmniESI (0.466) and EZSpecificity (0.463). On the \textbf{74 double-holdout splits}, where both the enzyme and substrate were unseen, it again ranked first (\textbf{0.436}, compared with 0.426 for the closest baseline; Table~\ref{tab:overall}). OmniESI and ProSmith had higher Pooled AP and Column AP in these aggregate averages, which compare scores across enzymes. Six of the panels have 16 splits each and hold out single proteins without clustering at $<40\%$ sequence identity. This allows baseline activity to transfer from close training homologs.

\begin{figure}[p]
\centering
\includegraphics[width=\linewidth,height=0.80\textheight,keepaspectratio]{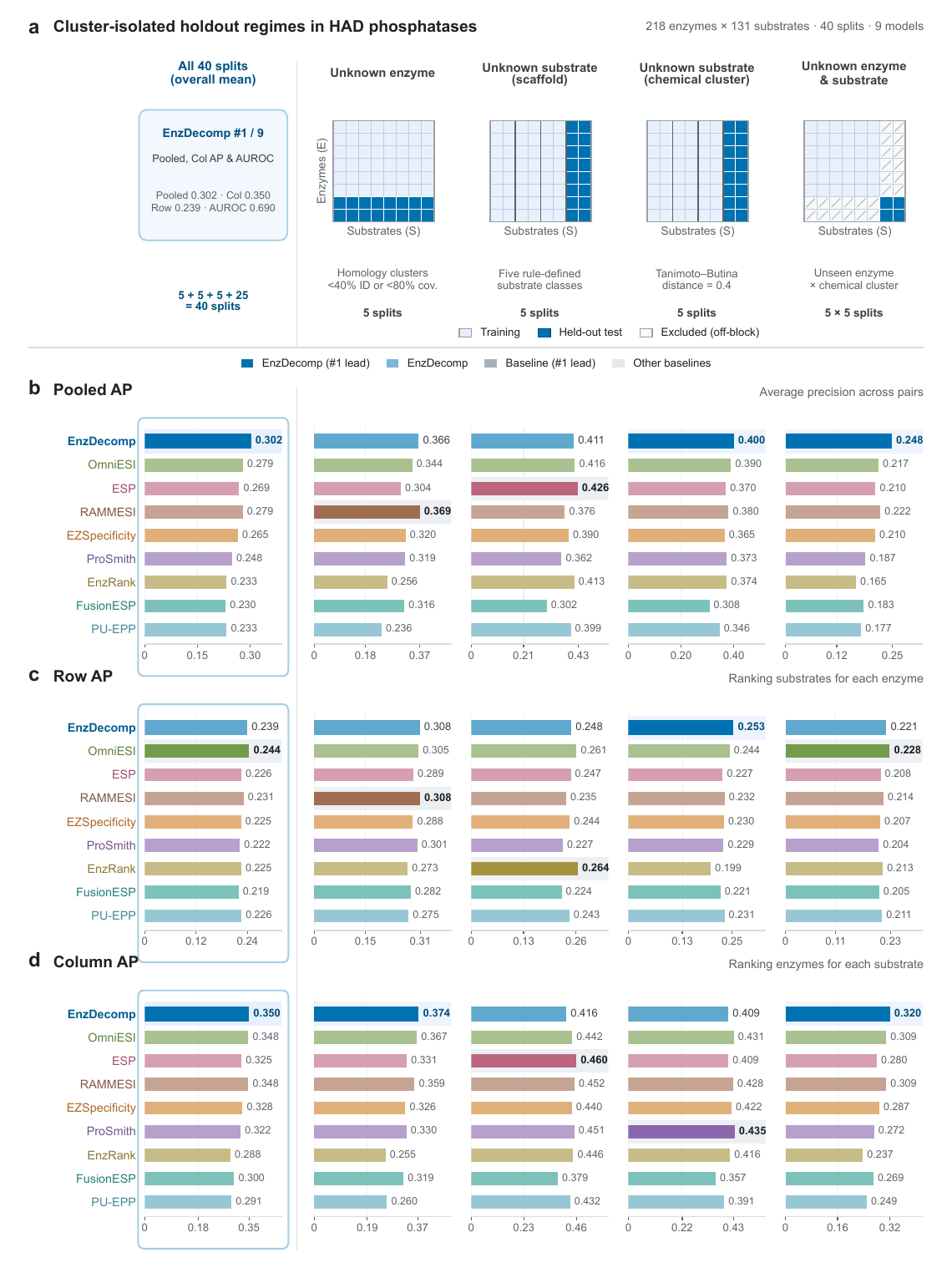}
\caption{\textbf{Extrapolating to unfamiliar enzyme clusters and chemical structures in HAD phosphatases.} \textbf{a}, The curated $218\times 131$ HAD phosphatase screen was split under four cluster-isolated regimes ($n=40$ splits in total). Five splits hold out MMseqs2 enzyme clusters ($<40\%$ identity or $<80\%$ coverage). Another five hold out biochemical substrate classes, and five hold out Butina clusters of Morgan fingerprints at Tanimoto distance $0.4$. The remaining 25 splits are $5\times 5$ double-cold splits in which both enzyme and chemical clusters are unseen. \textbf{b}--\textbf{d}, Equal-split mean whole-library Pooled AP (\textbf{b}), within-enzyme substrate ranking Row AP (\textbf{c}) and within-substrate enzyme retrieval Column AP (\textbf{d}) across EnzDecomp and eight retrained external baselines evaluated on identical held-out cells (all 1,080 individual split-level evaluations are plotted in Fig.~\ref{fig:s1}).}
\label{fig:had}
\end{figure}

The curated HAD phosphatase screen comprised $218\text{ enzymes}\times131\text{ substrates}$ and four holdout regimes (\textbf{40 splits} in total; Fig.~\ref{fig:had} and Fig.~\ref{fig:s1})\citep{huang2015}. HAD phosphatases share a Rossmannoid catalytic core with an $\mathrm{Asp}$ nucleophile. The C0, C1 and C2 cap domains differ between enzymes and control access to this core\citep{huang2015}. We examined generalization to enzyme clusters below 40\% sequence identity, to five biochemical substrate classes and to joint enzyme and chemical holdouts. A model trained on open-chain sugar phosphates, small phosphomonoesters and C--P phosphonates can thus encounter unfamiliar cap domains and substrates such as AMP, ATP and $\mathrm{NADP^+}$, cyclic hexose and disaccharide phosphates, pyrophosphates, or activated aryl phosphates such as \textit{p}-nitrophenyl phosphate and phosphotyrosine (Fig.~\ref{fig:had}a). Substrates in this HAD panel differed substantially in their susceptibility to phosphoryl-group cleavage. Phosphoanhydrides and phosphate esters with electron-withdrawing aryl leaving groups were hydrolyzed more readily across the superfamily than unactivated sugar phosphates or C--P phosphonates. We accounted for this superfamily-wide shared substrate reactivity through $C(s)$ when estimating matches $I(e,s)$ between new scaffolds and cap-domain pockets. Over the 40 splits, including those that excluded close homologs at the $<40\%$ identity threshold, EnzDecomp had the highest whole-library \textbf{Pooled AP} among the nine models (\textbf{0.302}; OmniESI, 0.279; retrained EZSpecificity, 0.265). It led on \textbf{Column AP (0.350)} and \textbf{AUROC (0.690)}. Within individual enzymes, Row AP was 0.239 for EnzDecomp and 0.244 for OmniESI. Across the four holdout regimes EnzDecomp exceeded 3D-docked EZSpecificity in \textbf{10 of the 12} AP comparisons (Fig.~\ref{fig:had}b--d and Table~\ref{tab:overall}).

\subsection*{Marginal freezing, family regimes and quantum-chemical cues}
\begin{figure}[p]
\centering
\includegraphics[width=\linewidth,height=0.73\textheight,keepaspectratio]{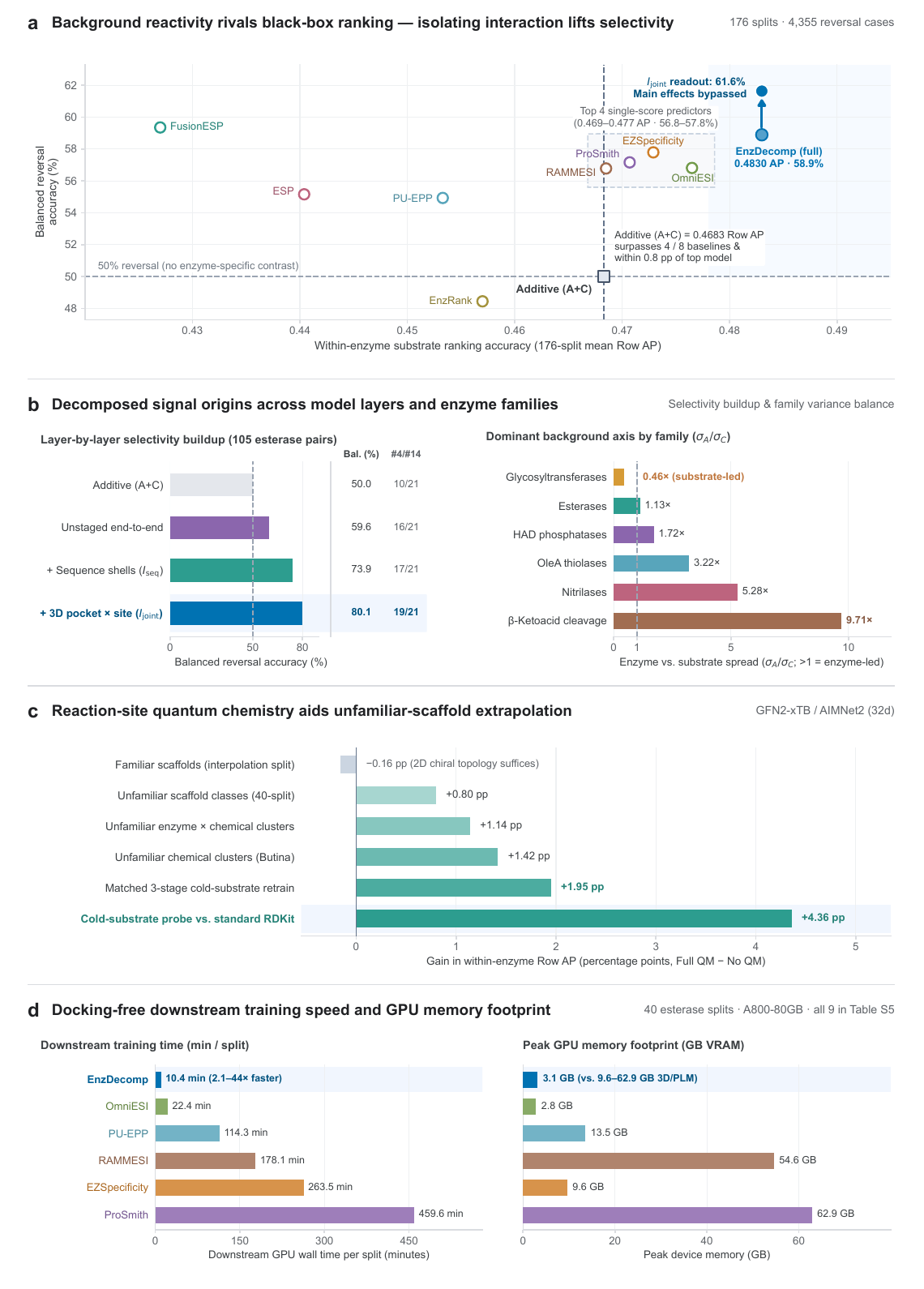}
\caption{\textbf{Opening the decomposed model: background ranking baselines, selectivity signal buildup and quantum-chemical descriptors.} \textbf{a}, The square shows what can be predicted from enzyme breadth and substrate reactivity alone ($A+C$). Over $n=176$ splits its Row AP is 0.468 (dashed vertical line), above four of the eight external scores and 0.8 percentage points below the highest. On $n=4,355$ reversal cases it gives 50.0\% balanced accuracy. From the same fit, the full score gives 0.483 Row AP (light-blue circle) and $I_{\rm joint}$ gives 61.6\% balanced accuracy (deep-blue circle). \textbf{b}, Left, balanced accuracy on the $n=105$ esterase substrate pairs from substrate holdouts as components are added. Parentheses give correct calls out of 21 for isomers \#4 and \#14. Right, ratio of enzyme-breadth and substrate-reactivity dispersions ($\sigma_A/\sigma_C$) in each of six families. \textbf{c}, Change in within-enzyme Row AP (percentage points) with 32 GFN2-xTB/AIMNet2 reaction-site descriptors on familiar and unfamiliar scaffolds, including comparison with 32 matched RDKit physicochemical descriptors. \textbf{d}, Downstream GPU wall time per split and peak allocated device memory on $n=40$ esterase splits (NVIDIA A800-SXM4-80GB). For all nine models, including the lightweight frozen-feature models, see Table~\ref{tab:runtime}.}
\label{fig:mechanism}
\end{figure}

After fitting enzyme breadth $A(e)$, shared substrate reactivity $C(s)$ and pocket interaction $I(e,s)$ separately, we tested whether post-hoc centering of an end-to-end network could replace the three-stage training procedure (Fig.~\ref{fig:mechanism}b--d). We trained the identical neural network from scratch without freezing marginal terms and tested it on 105 held-out esterase substrate pairs (Fig.~\ref{fig:mechanism}b, left, and Table~\ref{tab:internal_reversal}). Balanced reversal accuracy was \textbf{59.6\%} before centering and rose to \textbf{68.0\%} after double centering on the training reference panel, yet correct calls for the isomer pair \#4 and \#14 fell from 16/21 to 13/21. Freezing $C_r(s)$ and then $A(e)$ during training restricted the interaction layer to pocket residuals. Active-site sequence shells alone ($I_{\rm seq}$, pooled inner-shell $\leq 8$~\AA{} and outer-shell 8--15~\AA{} residues) reached \textbf{73.9\%} and 17/21, and attention between reaction sites and 3D reference-pocket blocks ($I_{\rm joint}$) reached \textbf{80.1\%} and \textbf{19/21}. The relative spread of fitted enzyme and substrate effects, $\sigma_A/\sigma_C$, varied by family (Fig.~\ref{fig:mechanism}b, right). The GT1 glycosyltransferases had the smallest ratio, $\mathbf{0.46\times}$, consistent with their sensitivity to acceptor nucleophilicity across the polyphenols and coumarins. The enzyme and substrate effects were closer in scale for broad-scope hydrolases, which vary in enzyme breadth and leaving-group lability (esterases, $\mathbf{1.13\times}$; HAD phosphatases, $\mathbf{1.72\times}$). Enzyme breadth dominated the narrower substrate series assayed for OleA thiolases, nitrilases and DUF849 $\beta$-ketoacid cleavage enzymes. These C--C or C--N cleavage families gave ratios of $\mathbf{3.22\text{--}9.71\times}$. In all six families the Stage~2 breadth $A(e)$ correlated with the estimate from sequence alone ($r=\mathbf{0.83\text{--}0.95}$). Strong pocket-specific effects reversed the family-wide substrate trend in \textbf{3--10\%} of enzyme--substrate pairs.

We retrained matched models with and without the 32 descriptors of the reaction site from GFN2-xTB and AIMNet2 (Fig.~\ref{fig:mechanism}c). For each protonation microstate we recorded the electronic energy, the partial atomic charge at the site and the polarization of the bond. The descriptors also include electrophilicity, nucleophilicity and frontier orbital Fukui indices. The local 2D chiral topology describes the acyl and alkoxy groups that recur in familiar esters. With site descriptors, the change in Row AP was $-0.16$ percentage points on these esters and \textbf{0.80 to 1.95 percentage points} on new scaffold classes or chemical clusters. On cold-substrate splits, Row AP was \textbf{4.36 percentage points} higher with the 32 quantum channels than with the 32 matched RDKit physicochemical descriptors (Fig.~\ref{fig:mechanism}c). Stage~1 used the site descriptors to fit the reactivity prior $C_r(s)$ and gained \textbf{+2.36 percentage points} on new functional groups. In Stage~3, the site vectors serve as electronic attention queries over the keys and values from the 3D pocket blocks.

Once a substrate has been encoded, its reaction-site descriptors can be used for all the enzymes in the screen. An enzyme's reference active-site pocket can likewise be used for all the substrates. This avoids $N_E\times N_S$ separate 3D docking calculations. On the 40-split esterase screen, five fits and their validation calibration cost \textbf{10.4~min per split} (2.09~min per run) and \textbf{3.1~GB} of peak GPU memory on NVIDIA A800 GPUs (Fig.~\ref{fig:mechanism}d and Table~\ref{tab:runtime}). OmniESI used 22.4~min and 2.8~GB, so EnzDecomp ran \textbf{2.1-fold faster} at a slightly higher memory cost. EZSpecificity, which uses 3D docking, used 263.5~min and 9.6~GB. Relative to that model EnzDecomp was \textbf{25-fold faster} with \textbf{3.1-fold lower memory}. For multimodal ProSmith (459.6~min, 62.9~GB), the differences were \textbf{44-fold faster} and \textbf{20-fold lower memory}.

\clearpage
\section*{Discussion}
In a library search for a starting enzyme or an easily cleaved linker, enzyme breadth $A(e)$ and shared substrate reactivity $C(s)$ are helpful. Across the 176 splits, whole-library pooled AP was 0.315 with $I_{\rm joint}$ alone versus 0.520 for the full score including calibrated $A(e)$ and $C(s)$ (Table~\ref{tab:internal}). For two orthogonal linkers in a hydrogel, prodrug or synthetic cascade, one enzyme must cleave $s$ alone and another must cleave $t$ alone. Every row of the full score carries the same shared substrate offset $C(s)-C(t)$. A pairwise comparison without that shared substrate difference recovered boundary preferences for constitutional isomers, epimers and enantiomers without docking. Post-hoc double-centering of raw scores on the training panel improved cross-panel balanced accuracy for all eight external models (Fig.~\ref{fig:selectivity}d). Post-hoc subtraction still fell short of staged training. On held-out esterase substrate pairs using the same backbone, post-hoc centering after end-to-end training gave 68.0\% balanced accuracy, compared with 80.1\% for staged training with centering during training (Table~\ref{tab:internal_reversal}).

For a candidate enzyme set, Readout~1 ($\hat f_{\rm EnzDecomp}$) provides a ranking in which to test compound--enzyme pairs on new sequences and scaffolds. From the same fit one could use Readout~2 ($I_{\rm joint}$) to select enzymes with opposite predicted preferences between constitutional isomers or stereoisomers. The spread of the fitted effects provides another description of the panel. Among glycosyltransferases, $\sigma_A/\sigma_C=\mathbf{0.46\times}$, so that the estimated substrate effects varied more than twice as much as enzyme breadth. The relative importance of acceptor and leaving-group chemistry here is consistent with the contribution from quantum descriptors for unfamiliar functional groups. In the enzyme-led cleavage families the ratio was $\mathbf{3.22\text{--}9.71\times}$. Changes in enzyme breadth, including variation in expression, could make substantial differences to the expected hit rate even before evaluating the specific active site.

Several limits apply to our experiments. Each panel records binary hits under one family's assay conditions, so score rankings reflect relative activity under those conditions, not $k_{\mathrm{cat}}/K_{\mathrm{M}}$ or $\Delta G^\ddagger$. Isomers were assayed separately as individual compounds, so evaluation measures selectivity between distinct molecules, not competing reaction sites within a multifunctional molecule. One fixed reference active-site pocket per enzyme makes downstream training after feature caching 25 times faster than pairwise docking. These fixed pockets, used only for dense single-family panels, miss binding-induced rearrangements of flexible cap domains or charged side chains. Small pocket ensembles sharing cached reaction-site descriptors could test flexibility. Without explicit 3D docking, local chirality has physical limits: $I_{\rm joint}$ achieves 55.1\% balanced accuracy across 17 cases with both enantiomers unseen, below EZSpecificity's raw 57.0\%, despite higher contrast AUROC (0.572 versus 0.546, Table~\ref{tab:isomers}). Two HAD phosphatase panels dominate the 4,355 cross-panel reversals (Methods). Post-hoc centered ProSmith exceeds $I_{\rm joint}$ on held-out esterase pairs (84.0\%, Table~\ref{tab:reversal}) and unseen constitutional isomers (76.0\%, Table~\ref{tab:isomers}), reflecting large-scale SMILES pretraining's advantage for 2D connectivity despite stereoisomer collapse. CLEAN\citep{clean2023}, CLIPZyme\citep{clipzyme2024} and Horizyn-1\citep{horizyn2026} narrow vast, sparse sequence databases to candidate families. Subsequent family screening, directed-evolution libraries and 3D affinity models including Boltz-2\citep{boltz2} still mix enzyme breadth with substrate lability.

\vspace{1.0em}
{\color{rulegray}\hrule height 0.6pt}
\vspace{0.2em}
{\small
\makeatletter
\renewcommand{\subsection}{\@startsection{subsection}{2}{\z@}%
  {-2.0ex \@plus -0.5ex \@minus -0.2ex}%
  {0.8ex \@plus 0.2ex}%
  {\normalfont\normalsize\bfseries\raggedright\hyphenpenalty=10000\exhyphenpenalty=10000\color{ink}}}
\makeatother
\section*{Methods}
\subsection*{Experimental screening panels and split construction}
In each experiment, enzymes $e\in E$ from an enzyme family were assayed against substrates $s\in S$ from that family's compound library. Each panel contains tens to hundreds of homologs tested under uniform experimental conditions. Assayed pairs with activity were recorded as $y_{es}=1$, and pairs assayed as inactive were recorded as $y_{es}=0$. We compiled six screening datasets from Goldman et al.\citep{goldman2022,goldmanDataset} and Cui et al.\citep{ezspecificity,ezDataset}, yielding eight reaction matrices and 176 training, validation and test splits. Matrix dimensions and isomer-pair counts for each panel are listed in Table~\ref{tab:panels}.

For six panels we used the partitions and binary activity thresholds distributed with EZSpecificity\citep{ezspecificity,ezDataset}. Each panel has four splits for each of four conditions ($4\times 4=16$). The test set holds out random pairs, enzymes, substrates, or both enzymes and substrates. Panel dimensions are given as enzymes by substrates. The metagenomic esterase panel contains $146\times 96$ entries (Mart{\'\i}nez-Mart{\'\i}nez et al.\citep{martinez2018}) and the haloacid dehalogenase (HAD) phosphatase panel $218\times 165$ (Huang et al.\citep{huang2015}). For the OleA thiolase superfamily, Robinson et al. measured whole-cell hydrolysis of activated \textit{p}-nitrophenyl esters ($73\times 15$)\citep{robinson2020}. The plant GT1 glycosyltransferase screen covers a range of polyphenol and coumarin acceptors ($55\times 104$, Yang et al.\citep{yang2018}). The remaining panels comprise nitrilases ($18\times 38$, Black et al.\citep{black2015}) and DUF849 $\beta$-ketoacid cleavage enzymes ($161\times 17$, Bastard et al.\citep{bastard2014}).

The published enzyme and substrate holdouts exclude an individual sequence or compound. In the two largest screens we could also exclude its close relatives. For esterases we took the complete $147\times 96$ activity matrix directly from Supplementary Tables~S1 and~S3 of Mart{\'\i}nez-Mart{\'\i}nez et al.\citep{martinez2018}, with assayed hydrolytic activity $>0$ scored as a positive hit. Five of the 40 splits held out enzyme homology clusters defined with MMseqs2\citep{mmseqs2} ($<40\%$ sequence identity or $<80\%$ mutual coverage). Five held out Bemis--Murcko chemical scaffolds\citep{bemismurcko}, using generic graphs that retained the topology of acyclic aliphatic esters. We made another five splits by Tanimoto--Butina clustering\citep{butina} of radius-2 Morgan fingerprints\citep{ecfp} at distance $0.4$. We crossed an enzyme cluster with a chemical cluster to produce $5\times 5=25$ more splits. For HAD phosphatases, all 131 substrates with unambiguous structures in Huang et al.\citep{huang2015} formed a $218\times 131$ panel. We made five MMseqs2 enzyme-cluster splits ($<40\%$ identity or $<80\%$ mutual coverage) and five substrate-class splits. The five classes were nucleosides and nucleotides (30 compounds), cyclic and open-chain sugar phosphates (17), pyro- and polyphosphates (4), aromatic and heterocyclic phosphates (4), and acyclic aliphatic phosphates and C--P phosphonates (76). We also made five Tanimoto--Butina splits using Morgan fingerprints at distance $0.4$, and $5\times 5=25$ splits holding out both enzyme and chemical clusters. Across all eight panels there were 24 random-pair splits and 152 extrapolation splits. Among the extrapolation splits, 74 held out both enzymes and substrates.

\subsection*{Candidate reaction-site indexing and quantum-chemical descriptors}
We standardized all substrates with RDKit\citep{rdkit}, retaining explicit tetrahedral stereochemistry. The assays report whole-molecule activity without atom-mapped reaction bonds or solution protonation states, so we enumerated candidate reactive groups and plausible protonation microstates for each molecule. Each candidate site in each microstate has 160 channels. The initial 128 local RDKit channels\citep{rdkit} describe 2D bond topology and local tetrahedral chiral parity, preserving stereochemical distinctions lost in bond graphs alone, canonical 1D SMILES embeddings or 2D descriptors compressed by principal-component analysis. The remaining 32 channels were calculated with semiempirical tight-binding GFN2-xTB\citep{gfn2xtb} and the AIMNet2 neural-network potential\citep{aimnet2}. For each microstate we recorded electronic energy, site partial atomic charge and bond polarization. The site channels add electrophilicity, nucleophilicity and frontier-orbital Fukui indices.

A circular fingerprint represents the part of the molecule outside the immediate reaction site as well as the site itself. For each split, we fitted the feature standardizer, linear input projection and channel gates using only training substrates $S_T$. For $|S_T|<20$, we retained only whole-molecule fingerprints and set local 2D and quantum channels to zero to prevent overfitting. Within each protonation microstate, we pooled candidate-site logits using temperature-normalized log-sum-exp and a learnable gated mean, then averaged across microstates to obtain one substrate prediction.

\subsection*{Docking-free reference active-site and protein sequence encoding}
An enzyme's fixed 3D reference active-site pocket can be used for all the substrates. This avoids $N_E\times N_S$ separate 3D docking calculations. For each structurally resolved enzyme, we encode one reference structure with ATOMICA\citep{atomica} for local 3D geometric interaction blocks and surrounding residues with ESM-C 600M\citep{esmc} (1,152 features per residue). We use no docked query-ligand coordinates. For the esterase panel and the four smaller panels, we partition residues and 3D pocket blocks around the reference active-site centroid into inner ($\leq 8$~\AA{}) and outer (8--15~\AA{}) shells. For the HAD phosphatase panel, we align the conserved Rossmannoid catalytic core surrounding the $\mathrm{Asp}$ nucleophile, variable C0, C1 and C2 cap domains, and hinge regions across homologs. For enzymes without a 3D reference structure, we set the ATOMICA pocket-block mask to zero, leaving only the active-site ESM-C sequence shells active.

\subsection*{Three-stage fit and training-panel centering}
An active pair in a family screen can arise from a broad-scope enzyme, a chemically labile substrate or a specific active-site match. We therefore fit enzyme breadth $A(e)$, shared substrate reactivity $C(s)$ and enzyme--substrate recognition $I(e,s)$\citep{lengerich2020} sequentially in three stages (Fig.~\ref{fig:architecture}c). Stage~1 fits the substrate term $C_r(s)$ on the experimentally observed training cells $M_T$ from whole-molecule substrate fingerprints plus chiral and quantum site descriptors, and the Stage~1 weights are then frozen. Stage~2 fits $A(e)$ as an offset above frozen $C_r(s)$ from pooled ESM-C active-site shells and ATOMICA reference-pocket blocks, with equal total binary cross-entropy weight per enzyme (including enzymes with few assayed training substrates), and the Stage~2 weights are then frozen. Stage~3 uses candidate site vectors as attention queries ($Q$) and 3D pocket features as keys and values ($K,V$). It combines the attended pocket context with pooled sequence shells and global pocket features, and applies soft-maximum and gated pooling over sites (Fig.~\ref{fig:architecture}c, \textbf{3a}) to give residual logits $D_{es}$ above $A(e)+C_r(s)$.

An interaction unrestricted in its mean can fit an offset for a whole row or column even after the first two stages have been frozen. We remove both offsets from $D_{es}$ during training by evaluating it throughout the reference rectangle $E_T\times S_T$ (Fig.~\ref{fig:architecture}c, \textbf{3b}),
\begin{equation}
 J_{es}=D_{es}-\langle D_{et}\rangle_{t\in S_T}
              -\langle D_{is}\rangle_{i\in E_T}
              +\langle D_{it}\rangle_{i\in E_T,t\in S_T},
 \label{eq:joint}
\end{equation}
where the means run over training substrates $S_T$ or training enzymes $E_T$ as indicated. Even an unmeasured cell contributes its predicted score to the mean on $E_T\times S_T$, giving $J_{es}$ exactly zero mean in each training row and column. Only the observed cells $M_T$ enter the loss, a combination of binary cross-entropy and the differentiable Smooth-AP ranking loss\citep{smoothap}. We used AdamW\citep{adamw} with a learning rate of $3\times 10^{-4}$. In Stage~2 $A(e)$ and Stage~3 $J_{es}$, the pocket blocks are masked if the structure is missing and have higher variance for distant homologs. A branch $S_{es}$ trained alongside these stages uses sequence and substrate features alone. We pooled ESM-C by shell or region and joined it with the pooled substrate descriptors. On $E_T\times S_T$ we computed $a^S_e=\langle S_{et}\rangle_{t\in S_T}$, $c^S_s=\langle S_{is}-a^S_i\rangle_{i\in E_T}$ and $i^S_{es}=S_{es}-a^S_e-c^S_s$. A sequence--chemistry prior ($\widetilde a^S_e$, $\widetilde c$ and $I_S$) also covers missing structures and distant reference homologs.

\subsection*{Task-matched readouts and support-aware validation calibration}
Given a pair of substrates, Readout~2 ($I_{\rm joint}$) compares their run-averaged centered interactions $\bar J_{es}$ from Stage~3 (Fig.~\ref{fig:architecture}d). No validation weights are estimated for this comparison. When plotting the components on a common scale, we multiply by the positive training-pair dispersion ratio $s(R)/s(J)>0$. Neither the sign nor the order of the contrasts changes. Readout~1 ($\hat f_{\rm EnzDecomp}$) is obtained from the same trained model by combining $\bar A$, the centered Stage~1 term $R$, the sequence priors $(\widetilde a^S_e,\widetilde c,\bar I_S)$ and $\bar I_J$. We align their variances on the training reference. Training substrate count and each split's validation set determine the support shrinkage $(w_M,w_V)$ without test labels (Supplementary Section~\ref{sec:cal}). Limited support draws auxiliary weights toward the additive anchor $A+C_r$. With sufficient support we maximize validation $\mathrm{RowAP}+\tfrac{1}{2}\mathrm{ColAP}+\tfrac{1}{4}\mathrm{DirectionAccuracy}$ with Row AP within $\tau=0.01$ of conservative anchor $B$. Screening accuracy was similar throughout when we chose the calibration family and $\tau$ by leaving one experimental source out. Across all 176 splits Row AP was 0.4829 versus 0.4830 (Table~\ref{tab:overall}).

\subsection*{External baseline retraining, post-hoc centering and quantum-descriptor probes}
We retrained eight published predictors from scratch on the same 176 training--validation splits and compared them with EnzDecomp on the same held-out test entries. The eight predictors are the multimodal language model ProSmith\citep{prosmith}, the conditional and retrieval-augmented fusion models OmniESI\citep{omniesi}, RAMMESI\citep{rammesi}, FusionESP\citep{fusionesp} and PU-EPP\citep{puepp}, ESP's gradient-boosted trees\citep{esp}, EnzRank's convolutional network\citep{enzrank} and the docked-complex 3D graph network EZSpecificity\citep{ezspecificity}. For EZSpecificity, we randomly initialized downstream 3D docked-complex graph network weights on each split, as ESIBank contains data overlapping these family screens\citep{ezspecificity}. On every split, training was repeated five times and predicted test-cell logits were averaged across the five runs before metric computation for both EnzDecomp and all external baselines (Tables~\ref{tab:overall}--\ref{tab:isomers}); symmetric unaveraged single-run means $\pm$ s.d.\ across individual runs are reported in Table~\ref{tab:fiverun}.

We double-centered each fitted model's scores $F$ using its own training reference panel without test labels, as in Eq.~\eqref{eq:joint}, $F^{\mathrm{cent}}_{es}=F_{es}-\langle F_{et}\rangle_{t\in S_T}-\langle F_{is}\rangle_{i\in E_T}+\langle F_{it}\rangle_{i\in E_T,t\in S_T}$ (Supplementary Section~\ref{sec:supp_selectivity}). On a complete reference panel, enzyme rank order and contrast AUROC remain the same for any substrate pair. We checked whether post-hoc centering of an end-to-end network can replace three-stage training (Tables~\ref{tab:internal} and~\ref{tab:internal_reversal}). We trained the identical neural network from scratch without freezing the marginal terms, and tested the staged uncentered residual without centering during training. On familiar and unfamiliar scaffold splits we retrained matched three-stage models with and without the 32 GFN2-xTB and AIMNet2 reaction-site channels. On cold-substrate splits we compared the 32 quantum reaction-site channels with 32 standard RDKit physicochemical descriptors\citep{gfn2xtb,aimnet2,rdkit}. We timed the downstream jobs and recorded peak GPU memory on NVIDIA A800-SXM4-80GB hosts over the 40 esterase splits (Table~\ref{tab:runtime}).

\subsection*{Screening metrics, stereoisomer classification and reversal statistics}
We scored held-out activity using the area under the receiver operating characteristic curve (AUROC) and average precision (AP)\citep{prroc}. \emph{Pooled AP} uses all observed test pairs. \emph{Row AP} ranks tested substrates within each enzyme and assigns zero to a row with assayed substrates but no active hits. \emph{Column AP} ranks test enzymes for each substrate. Column AP was defined on 175 of the 176 splits, including 151 of 152 non-random splits and 73 of 74 double-holdout splits. Each split had equal weight in the corresponding averages.

For a substrate pair $(s,t)$, we took test enzymes assayed on both compounds that were active on just one, with observed preference $z_e=y_{es}-y_{et}\in\{-1,+1\}$. We kept pair--split cases with $\geq 20$ discordant enzymes and $\geq 20\%$ minority preference. In all, across the 176 splits there are 4,355 high-support pair--split cases from 142 splits (3,127 distinct substrate pairs and 187,007 discordant enzyme--pair comparisons). They divide into 3,068 curated HAD, 815 distributed HAD, 336 workbook esterase, 105 distributed esterase, 21 glycosyltransferase and 10 DUF849 cases. Across the substrate-holdout splits, 105 esterase substrate pairs sampled on common support gave 3,472 comparisons (Table~\ref{tab:reversal}). No credit is given for an exact tie ($|f(e,s)-f(e,t)|\leq 10^{-12}$), which scores zero in both the $+1$ and $-1$ groups. On $\Delta f_e=f(e,s)-f(e,t)$, direction accuracy is the overall directional rate, balanced accuracy averages recall across the $+1$ and $-1$ groups, and cAUROC gives threshold-free contrast AUROC.

We grouped the 566 distinct RDKit-standardized compounds from the seven deduplicated libraries by molecular formula. Within each group we compared the connectivity and stereochemistry layers of standard InChIKeys. Distinct connectivity gave 594 constitutional isomer pairs. For compounds with identical connectivity, partial inversion of tetrahedral stereocenters gave 233 diastereomer/epimer pairs, and complete inversion gave 43 mirror-image pairs. The 870 distinct pairs occurred 907 times across the panels (Table~\ref{tab:panels}). We measured pairwise $L_2$ distances between ChemBERTa-2 SMILES embeddings. For the $218\times 131$ HAD library, we measured the fraction of squared distance retained after compressing the 2D descriptors to 128 principal components. We called an isomer comparison cold when $s\notin S_T$ or $t\notin S_T$. This definition gave 780 high-support cases, comprising 326 distinct pairs and 29,395 discordant comparisons. Of these cases, 508 involved constitutional isomers, 237 involved diastereomers or epimers and 35 involved enantiomers. Both compounds were held out ($s\notin S_T$ and $t\notin S_T$) in 17 enantiomer cases covering 9 distinct pairs (Table~\ref{tab:isomers}).

\section*{Data availability}
The reaction matrices are publicly available from primary wet-lab screens. Mart{\'\i}nez-Mart{\'\i}nez et al.\ reported metagenomic esterases\citep{martinez2018}, Huang et al.\ HAD phosphatases\citep{huang2015}, Robinson et al.\ OleA thiolases\citep{robinson2020}, Yang et al.\ GT1 glycosyltransferases\citep{yang2018}, Black et al.\ nitrilases\citep{black2015}, and Bastard et al.\ DUF849 $\beta$-ketoacid cleavage enzymes\citep{bastard2014}. The eight panels were curated via Goldman et al.\citep{goldman2022,goldmanDataset} and Cui et al.\citep{ezspecificity,ezDataset}. Table~\ref{tab:panels} gives the dimensions of the reaction matrix and the number of isomers for each panel. Standardized substrate structures, InChIKey stereoisomer classifications for all 870 unique isomer pairs, reference active-site pocket coordinates, cached descriptors and evaluation tables for all 176 train--validation--test splits will be made publicly available upon publication and can be provided to editors and reviewers upon request during peer review.

\section*{Code availability}
The EnzDecomp source code covers candidate reaction-site chiral and GFN2-xTB/AIMNet2 descriptor indexing, reference-pocket and active-site sequence encoding, three-stage fitting with training-panel double centering, and validation calibration. The source code, model checkpoints and deterministic scripts reproducing the main-text and Supplementary Information figures and tables will be made publicly available upon publication, and can be provided to editors and reviewers upon request during peer review.

\section*{Acknowledgments}
We thank Jiaxiang Zhang (School of Medicine and Pharmacy, Ocean University of China) for helpful discussions on experimental assay controls, negative-entry curation, and isomer selectivity in family-wide screens. We also thank the Shanghai Artificial Intelligence Laboratory for providing computational resources.

\section*{Author contributions}
M.X. conceived the study, curated the experimental screening matrices and isomer benchmarks, designed and implemented the EnzDecomp framework and feature-encoding pipeline, performed the benchmark experiments and statistical analyses, prepared the initial figures, and drafted the manuscript. X.Q. contributed to algorithmic refinement, figure preparation, and manuscript revision. Both authors reviewed and approved the final manuscript.

\section*{Competing interests}
The authors declare no competing interests.

\bibliographystyle{unsrtnat}
\bibliography{references}
}

\appendix
\clearpage
\setcounter{table}{0}
\renewcommand{\thetable}{S\arabic{table}}
\renewcommand{\theHtable}{supp.\arabic{table}}
\setcounter{figure}{0}
\renewcommand{\thefigure}{S\arabic{figure}}
\renewcommand{\theHfigure}{supp.\arabic{figure}}

{\raggedright\hyphenpenalty=10000\exhyphenpenalty=10000
 {\LARGE\bfseries\color{ink}EnzDecomp: Disentangling background reactivity from active-site selectivity in enzyme screens\par}
 \vspace{0.35em}
 {\large\bfseries\color{ink}Supplementary Information\par}
}\vspace{0.4em}
\vspace{0.6em}

\noindent Section~\ref{sec:supp_data} details the screening reaction matrices, stereoisomer statistics and retraining protocols for the external baselines, and Section~\ref{sec:cal} gives the variance-alignment and support-shrinkage equations together with screening result tables across all holdout regimes. Preference reversals, post-hoc double-centering and cold stereoisomer tests are presented in Section~\ref{sec:supp_selectivity}. Section~\ref{sec:supp_ablations} summarizes GPU runtimes, ablation experiments and individual-run stability. The contents of the reproducibility archive, which will be made publicly available upon publication, are listed in Section~\ref{sec:repro}.

\section{Benchmark construction, reaction-matrix inventory and baseline protocols}\label{sec:supp_data}

\subsection{Reaction matrices and stereoisomers}
\paragraph{Experimental matrices.}
In each experiment an enzyme $e\in E$ was tested on substrates $s\in S$ from its family library. The screens covered tens to hundreds of homologs and used common conditions within a panel (Table~\ref{tab:panels})\citep{goldman2022,goldmanDataset}. We used $y_{es}=1$ for an assayed active pair and $y_{es}=0$ for an assayed inactive pair. For six panels we used the 16 splits distributed with EZSpecificity\citep{ezspecificity,ezDataset}. Each panel has four splits for each of four conditions ($4\times 4=16$). The test set holds out random pairs, enzymes, substrates, or both enzymes and substrates. In the two largest screens we also built two 40-split panels directly from primary records. For esterases we took the complete $147\times 96$ matrix from Mart{\'\i}nez-Mart{\'\i}nez et al.\citep{martinez2018}. For haloacid dehalogenase (HAD) phosphatases, all 131 substrates with unambiguous structures in Huang et al.\citep{huang2015} formed a $218\times 131$ panel. In each 40-split panel, five splits hold out MMseqs2 sequence-homology clusters ($<40\%$ sequence identity or $<80\%$ mutual coverage\citep{mmseqs2}). Five more hold out biochemical substrate classes or Bemis--Murcko scaffolds\citep{bemismurcko}, five hold out Tanimoto--Butina Morgan-fingerprint clusters at distance $0.4$\citep{butina,ecfp} and $5\times 5=25$ hold out enzyme clusters and chemical clusters simultaneously (\textbf{176 splits} in total).

\paragraph{Isomer and enantiomer census.}
After standardizing the substrates in RDKit\citep{rdkit} with explicit tetrahedral stereochemistry retained, we compared only compounds from the same library with the same formula. A difference in the connectivity layer of the standard InChIKey defines a constitutional-isomer pair. For identical connectivity, we compared the stereochemistry layers: complete inversion at all tetrahedral centers defines a mirror-image enantiomer pair, and partial inversion a diastereomer/epimer pair. The seven deduplicated substrate libraries contained 566 unique standardized substrates and \textbf{870 unique isomer pairs}: \textbf{594} constitutional, \textbf{233} diastereomer/epimer and \textbf{43} enantiomer pairs. The same census by panel gives 907 occurrences over eight panels (Table~\ref{tab:panels}).

\begin{table}[!htbp]
\centering
\caption{\textbf{Experimental origins, dimensions and isomer pair counts of the eight screening panels.} Isomer pair counts sum constitutional, diastereomeric/epimeric and enantiomeric substrate pairs within each library (870 unique isomer pairs across the seven deduplicated substrate libraries; 907 across all eight panels).}
\label{tab:panels}
\small
\begin{tabular}{@{}lrrrrl@{}}
\toprule
Panel & Enzymes & Substrates & Isomer pairs & Splits & Primary screen \\
\midrule
Esterase, distributed & 146 & 96 & 37 & 16 & Ref.~\citenum{martinez2018} \\
Esterase, source workbook & 147 & 96 & 37 & 40 & Ref.~\citenum{martinez2018} \\
HAD phosphatase, distributed & 218 & 165 & 495 & 16 & Ref.~\citenum{huang2015} \\
HAD phosphatase, curated & 218 & 131 & 312 & 40 & Ref.~\citenum{huang2015} \\
OleA thiolase & 73 & 15 & 0 & 16 & Ref.~\citenum{robinson2020} \\
Glycosyltransferase acceptors & 55 & 104 & 13 & 16 & Ref.~\citenum{yang2018} \\
Nitrilase & 18 & 38 & 12 & 16 & Ref.~\citenum{black2015} \\
DUF849 ($\beta$-ketoacid cleavage enzymes) & 161 & 17 & 1 & 16 & Ref.~\citenum{bastard2014} \\
\bottomrule
\end{tabular}
\end{table}

\subsection{External baselines}
\paragraph{Retraining.}
We trained all eight external models from scratch on the same 176 train--validation splits as EnzDecomp and scored them on the same held-out test cells. These models were ProSmith\citep{prosmith}, OmniESI\citep{omniesi}, EZSpecificity\citep{ezspecificity}, RAMMESI\citep{rammesi}, ESP\citep{esp}, FusionESP\citep{fusionesp}, PU-EPP\citep{puepp} and EnzRank\citep{enzrank}. ESIBank contains reactions from the enzyme families studied here\citep{ezspecificity}, so we omitted EZSpecificity's pretrained ESIBank weights and randomly initialized its downstream complex-graph weights on each split. For EnzDecomp, the feature standardizers, linear projections, channel gates and $E_T\times S_T$ double-centered means for each split were fitted solely on its training cells, without access to validation or test labels. On every split, training was repeated five times and predicted test-cell logits were averaged across runs prior to metric computation for both EnzDecomp and all eight external baselines (Tables~\ref{tab:overall}--\ref{tab:isomers}); unaveraged single-run means $\pm$ s.d.\ are reported alongside in Table~\ref{tab:fiverun}.

\section{Calibration and screening tables}\label{sec:cal}

\subsection{Variance alignment and shrinkage}
Without sequence priors or post-hoc calibration, the three-stage model $A+C_r+\gamma J$ achieved \textbf{0.4781 Row AP} across 176 splits, exceeding every external model (Table~\ref{tab:internal}). Readout~2 ($I_{\rm joint}$) reads the centered interaction term directly, with \emph{zero} post-processing. Sequence--chemistry priors cover missing, sparse or distantly homologous pocket structures. To construct Readout~1 ($\hat f_{\rm EnzDecomp}$) for screening, we aligned the 3D and sequence terms to the same variance scale over $E_T\times S_T$ and shrank their weights by validation support. Low validation support shifted the weights toward the additive baseline $A+C_r$. All calibration quantities used only each split's $E_T\times S_T$, observed training cells $M_T$ and validation cells, never test labels.

\paragraph{Reference anchor and empirical support factors.}
For each enzyme, $b_e=\langle C_{r,et}\rangle_{t\in S_T}$ is the mean Stage~1 prior over the training substrates. Subtracting it gives $R_{es}=C_{r,es}-b_e$. Here an overbar denotes a stage output averaged across runs (Methods), and $\operatorname{sd}$ is the population standard deviation. From the observed training pairs $M_T$, we take $s(X)=\operatorname{sd}_{(e,s)\in M_T}(\bar X_{es})+10^{-8}$. We put the auxiliary terms on the scale of $R$ with $\widetilde c=c^S s(R)/s(c^S)$, $I_S=i^S s(R)/s(i^S)$ and $I_J=J\,s(R)/s(J)$. The blended substrate prior and reference score for weights $(\alpha,\beta_0,\gamma_0)$ are
\begin{equation}
 C_{\mathrm{mix}}=\overline{b+(1-\alpha)R+\alpha\widetilde c},\qquad
 B=\bar A+C_{\mathrm{mix}}+\beta_0\bar I_S+\gamma_0\bar I_J,
\end{equation}
which gives $A+C_r$ when $(\alpha,\beta_0,\gamma_0)=(0,0,0)$. Here $n_E$ is the count of validation enzymes with both activity classes on at least two assayed substrates; $n_T=|S_T|$ is the number of training substrates and $n_V$ is the number of validation substrates. Set $w_M=\min(1,n_E/70)^2\min(1,n_T/30)$, $w_V=\min(1,\max(0,n_V-7)/25)$ and $p=w_M(0.35,0.20,0.20)$, and score $(1-w_V)p+w_Vw_Mr$ on the validation cells over $r\in\{0,0.20,0.35,0.50\}\times\{0,0.15,0.25,0.40\}^2$. At $w_V=0$ we need only score $p$. We keep the best candidate if $n_E\geq 3$, $n_V\geq 2$ and its validation Row AP exceeds that of $A+C_r$ by more than $10^{-6}$. Otherwise we use $(\alpha,\beta_0,\gamma_0)=(0,0,0)$.

\subsection{Sequence-branch alignment and dual-axis dispersion normalization}
To align the sequence breadth $\overline{a^S_e}$ with the Stage~2 enzyme breadth $A(e)$, we weight each enzyme by the number of observed training cells for that enzyme. From the Stage~2 scores averaged over training runs and $\overline{a^S_e}$, we compute the weighted means $m_A,m_{AS}$ and the weighted population standard deviations $s_A,s_{AS}$, with $10^{-8}$ added to each standard deviation. We then define $\widetilde a^S_e=m_A+(s_A/s_{AS})(\overline{a^S_e}-m_{AS})$ and score
\begin{equation}
 D^*_{es}=(1-a)\bar A_{es}+a\widetilde a^S_e+C_{\mathrm{mix},es}+\beta\bar I_{S,es}+\gamma\bar I_{J,es}
\end{equation}
over $a\in\{0,0.25,0.5,1\}$, $\beta\in\{0,\beta_0,0.25,0.5,1\}$ and $\gamma\in\{0,\gamma_0,0.25,0.5,1\}$. We use $B$ as an anchor and add the best admissible $D^*$ if the scores differ. For each anchor $D\in\{B,D^*\}$, we write the interaction weights as $(w_S,w_J)$. To estimate the dispersion for each substrate column, we take the mean $\mu_{Ls}$ and population standard deviation $d_{Ls}$ of $\bar I_{L,es}$ across $e\in E_T$ for $L\in\{S,J\}$. We use $\epsilon=\max(0.01\,s(R),10^{-8})$ in calculating the median column dispersion $t_L=\operatorname{median}_{s\in S_T}d_{Ls}+\epsilon$ and the bounded column multiplier $g_{Ls}=\operatorname{clip}(t_L/(d_{Ls}+\epsilon),0.25,4)$. A separate multiplier $u_A=\operatorname{clip}(\sqrt{s_A/s(R)},0.5,4)$ accounts for the relative variation across enzymes and substrates. The adjusted scores are
\begin{align}
 H_{L,es} &= (\bar I_{L,es}-c\mu_{Ls})(1-n+ng_{Ls}), &
 H^*_{L,es} &= H_{L,es}-(1-\eta)\langle H_{L,et}\rangle_{t\in Q}, \\
 Z_{es} &= D_{es}+\sum_{L\in\{S,J\}} w_L\left[(1-v+vu_A)H^*_{L,es}-\bar I_{L,es}\right],
\end{align}
for $n\in\{0.5,1\}$ and $v,\eta,c\in\{0,1\}$. The mean over $Q$ uses predictions across the query substrate block and requires no labels. On each split we maximize validation $\mathrm{RowAP}+\tfrac{1}{2}\mathrm{ColAP}+\tfrac{1}{4}\mathrm{DirectionAccuracy}$ with Row AP within $\tau=0.01$ of $\mathrm{RowAP}_{\mathrm{val}}(B)$. We also selected the calibration family and $\tau$ with one experimental source excluded at a time. The leave-one-source-out and default rules gave \textbf{0.4829 versus 0.4830 Row AP} across all 176 splits, with similar values in the other blocks of Tables~\ref{tab:overall} and~\ref{tab:internal}.

\subsection{Whole-library and per-axis screening results across holdout regimes}
\begin{figure}[p]
\centering
\includegraphics[width=\linewidth,height=0.76\textheight,keepaspectratio]{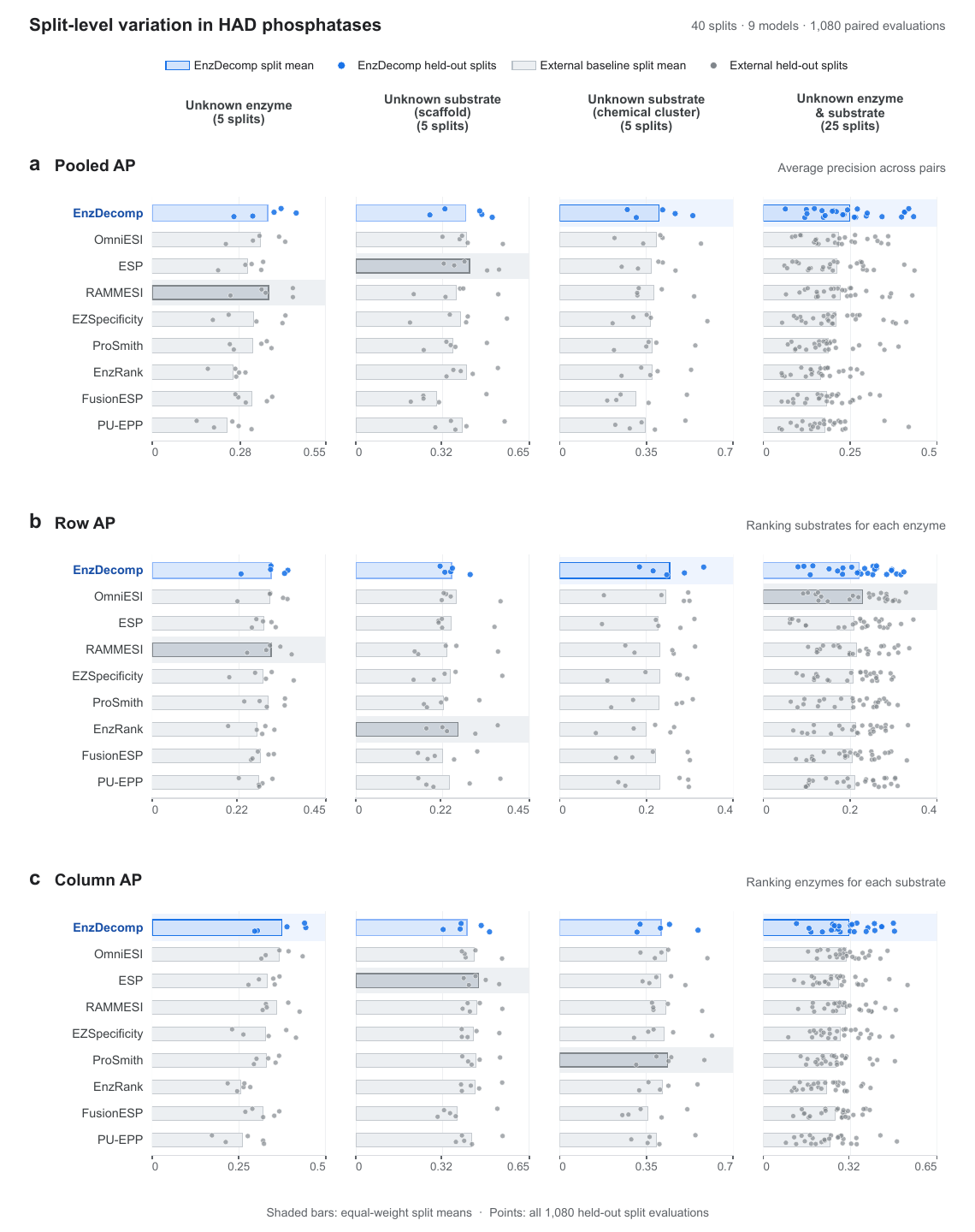}
\caption{\textbf{Split-level variation across all nine models on the 40-split HAD phosphatase benchmark.} Bars denote equal-weight split means across $n=40$ splits (evaluation protocol in Methods) and points display all $n=1,080$ paired held-out split evaluations across pooled AP (\textbf{a}), row AP (\textbf{b}) and column AP (\textbf{c}), corresponding to Fig.~\ref{fig:had}b--d and Table~\ref{tab:overall}.}
\label{fig:s1}
\end{figure}

Table~\ref{tab:overall} gives the means for all 176 splits, the 152 non-random splits, the 74 double holdouts and the 40 HAD phosphatase splits with cluster holdouts. For the curated HAD experiment, Fig.~\ref{fig:s1} shows \textbf{1,080 paired split-level evaluations} ($40\text{ splits}\times 9\text{ models}\times 3\text{ AP axes}$). EnzDecomp had the highest within-enzyme Row AP across 176 splits (\textbf{0.4830}), across the 152 OOD splits (\textbf{0.4747}) and across the 74 double holdouts (\textbf{0.4358}). On the 40-split HAD panel with $<40\%$ sequence-identity clustering, it had the highest whole-library Pooled AP (\textbf{0.3024}), Column AP (\textbf{0.3501}) and AUROC (\textbf{0.6901}).

\begin{table}[!htbp]\centering\scriptsize\setlength{\tabcolsep}{4.2pt}\renewcommand{\arraystretch}{0.96}
\caption{\textbf{Whole-library and per-axis activity ranking across holdout regimes.} Symmetric comparison on identical observed test cells using run-averaged test score matrices for all external baselines and EnzDecomp across all 176 splits, all 152 non-random out-of-distribution splits, all 74 double-holdout splits (where both enzymes and substrates are held out) and all 40 curated HAD phosphatase splits (Methods; symmetric single-run means $\pm$ s.d.\ across individual runs are reported in Table~\ref{tab:fiverun}). EnzDecomp is shown under both source-excluded leave-one-source-out (LOSO) combination-family selection and default validation selection. Column AP and AUROC are defined on 175 of 176 splits (151 of 152 non-random splits and 73 of 74 double-holdout splits). Bold marks the highest value within each block and column.}\label{tab:overall}
\begin{tabular}{@{}lrrrrr@{}}\toprule
Model / Readout & Splits & Row AP & Column AP & Pooled AP & AUROC\\ \midrule
\multicolumn{6}{@{}l}{\textit{Full benchmark across all eight panels (176 splits; 24 random + 152 out-of-distribution)}}\\
FusionESP\citep{fusionesp} & 176 & 0.4270 & 0.4995 & 0.4552 & 0.7314\\
ESP\citep{esp} & 176 & 0.4404 & 0.5200 & 0.4844 & 0.7507\\
PU-EPP\citep{puepp} & 176 & 0.4533 & 0.4911 & 0.4634 & 0.7403\\
EnzRank\citep{enzrank} & 176 & 0.4570 & 0.4889 & 0.4780 & 0.7573\\
RAMMESI\citep{rammesi} & 176 & 0.4685 & 0.5493 & 0.5221 & 0.7829\\
ProSmith\citep{prosmith} & 176 & 0.4707 & 0.5602 & 0.5303 & 0.7846\\
EZSpecificity\citep{ezspecificity} & 176 & 0.4729 & 0.5445 & 0.5269 & 0.7797\\
OmniESI\citep{omniesi} & 176 & 0.4765 & \textbf{0.5664} & \textbf{0.5390} & \textbf{0.7944}\\
EnzDecomp (LOSO rule selection) & 176 & 0.4829 & 0.5449 & 0.5188 & 0.7782\\
\textbf{EnzDecomp (ours)} & 176 & \textbf{0.4830} & 0.5453 & 0.5202 & 0.7782\\
\midrule
\multicolumn{6}{@{}l}{\textit{All non-random out-of-distribution splits across eight panels (152 splits)}}\\
FusionESP\citep{fusionesp} & 152 & 0.4144 & 0.4613 & 0.4192 & 0.7132\\
ESP\citep{esp} & 152 & 0.4283 & 0.4833 & 0.4469 & 0.7273\\
PU-EPP\citep{puepp} & 152 & 0.4425 & 0.4495 & 0.4230 & 0.7145\\
EnzRank\citep{enzrank} & 152 & 0.4518 & 0.4522 & 0.4478 & 0.7373\\
RAMMESI\citep{rammesi} & 152 & 0.4568 & 0.5122 & 0.4850 & 0.7616\\
ProSmith\citep{prosmith} & 152 & 0.4600 & 0.5223 & 0.4882 & 0.7626\\
EZSpecificity\citep{ezspecificity} & 152 & 0.4627 & 0.5063 & 0.4854 & 0.7573\\
OmniESI\citep{omniesi} & 152 & 0.4662 & \textbf{0.5269} & \textbf{0.4967} & \textbf{0.7738}\\
EnzDecomp (LOSO rule selection) & 152 & 0.4746 & 0.5096 & 0.4816 & 0.7605\\
\textbf{EnzDecomp (ours)} & 152 & \textbf{0.4747} & 0.5099 & 0.4819 & 0.7603\\
\midrule
\multicolumn{6}{@{}l}{\textit{All double-holdout splits across eight panels (74 splits; unseen enzymes and substrates)}}\\
FusionESP\citep{fusionesp} & 74 & 0.3725 & 0.3941 & 0.3260 & 0.6366\\
ESP\citep{esp} & 74 & 0.3758 & 0.4032 & 0.3355 & 0.6436\\
PU-EPP\citep{puepp} & 74 & 0.3966 & 0.3691 & 0.3181 & 0.6224\\
RAMMESI\citep{rammesi} & 74 & 0.4052 & 0.4382 & 0.3848 & 0.6915\\
ProSmith\citep{prosmith} & 74 & 0.4176 & 0.4401 & 0.3867 & 0.6929\\
OmniESI\citep{omniesi} & 74 & 0.4178 & \textbf{0.4455} & 0.3899 & \textbf{0.7070}\\
EZSpecificity\citep{ezspecificity} & 74 & 0.4211 & 0.4297 & \textbf{0.3901} & 0.6903\\
EnzRank\citep{enzrank} & 74 & 0.4261 & 0.3668 & 0.3594 & 0.6607\\
EnzDecomp (LOSO rule selection) & 74 & \textbf{0.4358} & 0.4298 & 0.3849 & 0.6964\\
\textbf{EnzDecomp (ours)} & 74 & \textbf{0.4358} & 0.4298 & 0.3849 & 0.6964\\
\midrule
\multicolumn{6}{@{}l}{\textit{Curated HAD phosphatase benchmark across all four holdout regimes (40 splits)}}\\
FusionESP\citep{fusionesp} & 40 & 0.2191 & 0.3002 & 0.2301 & 0.6355\\
ProSmith\citep{prosmith} & 40 & 0.2222 & 0.3218 & 0.2485 & 0.6542\\
EZSpecificity\citep{ezspecificity} & 40 & 0.2248 & 0.3282 & 0.2655 & 0.6608\\
EnzRank\citep{enzrank} & 40 & 0.2253 & 0.2878 & 0.2333 & 0.6338\\
ESP\citep{esp} & 40 & 0.2256 & 0.3249 & 0.2689 & 0.6600\\
PU-EPP\citep{puepp} & 40 & 0.2257 & 0.2913 & 0.2332 & 0.6150\\
RAMMESI\citep{rammesi} & 40 & 0.2310 & 0.3477 & 0.2792 & 0.6741\\
OmniESI\citep{omniesi} & 40 & \textbf{0.2439} & 0.3479 & 0.2794 & 0.6886\\
EnzDecomp (LOSO rule selection) & 40 & 0.2394 & \textbf{0.3501} & \textbf{0.3024} & \textbf{0.6901}\\
\textbf{EnzDecomp (ours)} & 40 & 0.2394 & \textbf{0.3501} & \textbf{0.3024} & \textbf{0.6901}\\
\bottomrule\end{tabular}\end{table}

\clearpage
\section{Selectivity reversals, post-hoc centering and stereoisomer diagnostics}\label{sec:supp_selectivity}

\subsection{Preference reversals and post-hoc centering}
\paragraph{Reversal cases.}
For a substrate pair $(s,t)$, we took test enzymes assayed on both compounds that were active on just one of the two, with observed preference $z_e=y_{es}-y_{et}\in\{-1,+1\}$. We kept pair--split cases with $\geq 20$ discordant enzymes and $\geq 20\%$ minority preference. In all, there are \textbf{4,355 high-support pair--split cases} from 142 splits (3,127 distinct substrate pairs and 187,007 discordant enzyme--pair comparisons). Across the substrate-holdout splits, \textbf{105 esterase substrate pairs} sampled on common support gave 3,472 comparisons (Table~\ref{tab:reversal}). No credit is given for an exact tie ($|f(e,s)-f(e,t)|\leq 10^{-12}$), which scores zero in both the $+1$ and $-1$ groups. On $\Delta f_e=f(e,s)-f(e,t)$, direction accuracy is the overall directional rate, balanced accuracy averages recall across the $+1$ and $-1$ groups, and cAUROC gives threshold-free contrast AUROC.

\paragraph{Post-hoc double centering.}
In an external score matrix $F$, a shared substrate offset can favor one substrate over another across all enzymes. We estimated each fitted model's background on its own training reference panel, using no test labels, and applied the double centering of Eq.~\eqref{eq:joint},
\begin{equation}
 F^{\mathrm{cent}}_{es} = F_{es} - \langle F_{et}\rangle_{t\in S_T} - \langle F_{is}\rangle_{i\in E_T} + \langle F_{it}\rangle_{i\in E_T,t\in S_T},
\end{equation}
where $E_T\times S_T$ is the split's training reference panel of enzymes and substrates. For a given substrate pair $(s,t)$, the row mean cancels in $F(e,s)-F(e,t)$, and removing $\langle F_{is}\rangle_{i\in E_T}-\langle F_{it}\rangle_{i\in E_T}$ subtracts the same constant from $F(e,s)-F(e,t)$ for every test enzyme $e$ when the reference panel is complete. Its sign relative to zero, and hence its agreement with the observed preference, can change, improving direction and balanced accuracy. Its rank among enzymes and the contrast AUROC remain the same. All \textbf{eight out of eight external models} improved in cross-panel balanced reversal accuracy in this test (Table~\ref{tab:reversal}, bottom). ProSmith went from 57.16\% to 60.44\% and OmniESI from 56.81\% to 60.03\%. The highest overall value was \textbf{61.63\%}, obtained by reading EnzDecomp's Stage~3 $I_{\rm joint}$ directly without post-processing.

\begin{table}[!htbp]\centering\scriptsize\setlength{\tabcolsep}{3.2pt}\renewcommand{\arraystretch}{0.96}
\caption{\textbf{Direction of preference among discordant enzymes across 4,355 cross-panel cases and 105 esterase substrate pairs.} The left columns cover 4,355 high-support pair--split cases from 142 splits (3,127 distinct substrate pairs and 187,007 discordant enzyme--pair comparisons). The right columns cover 105 esterase substrate pairs sampled on common support across four substrate holdouts (3,472 comparisons). For the example of held-out ethyl hexanoate (\#4) and hexyl acetate (\#14), the count is over the $n=21$ discordant esterases. All readouts and external baselines use run-averaged test score matrices (Methods). Results scored separately by run are given as mean $\pm$ s.d.\ in Table~\ref{tab:fiverun}. No credit is given for an exact tie. Across 4,355 cases raw tie rates were 4.1\% (ESP), 3.1\% (ProSmith), 3.1\% (EnzRank), 2.2\% (OmniESI) and 0.0\% (EnzDecomp). Dir and Bal give equal-case mean direction and balanced accuracy (\%), and cAUROC gives threshold-free contrast AUROC. Bold marks the best value per column in each block.}\label{tab:reversal}
\resizebox{\linewidth}{!}{\begin{tabular}{@{}lrrrrrrr@{}}\toprule
& \multicolumn{3}{c}{All cross-panel cases (4,355; 3,127 pairs)} & \multicolumn{4}{c}{Esterase substrate holdout (105 pairs)}\\ \cmidrule(lr){2-4}\cmidrule(l){5-8}
Model / Readout & Dir (\%) & Bal (\%) & cAUROC & Dir (\%) & Bal (\%) & cAUROC & \#4 vs.\ \#14\\ \midrule
\multicolumn{8}{@{}l}{\textit{Raw model readouts}}\\
EnzRank\citep{enzrank} & 50.81 & 48.45 & 0.5024 & 52.79 & 50.00 & 0.5238 & $10\,/\,21$\\
PU-EPP\citep{puepp} & 56.27 & 54.93 & 0.5980 & 75.30 & 72.29 & 0.8697 & $15\,/\,21$\\
ESP\citep{esp} & 57.51 & 55.16 & 0.6245 & 69.57 & 67.93 & 0.8396 & $14\,/\,21$\\
RAMMESI\citep{rammesi} & 58.36 & 56.78 & 0.6244 & 75.85 & 71.76 & 0.8581 & $11\,/\,21$\\
OmniESI\citep{omniesi} & 58.66 & 56.81 & 0.6267 & 79.39 & 77.67 & 0.8895 & $16\,/\,21$\\
ProSmith\citep{prosmith} & 58.48 & 57.16 & 0.6472 & \textbf{81.94} & 80.03 & \textbf{0.9147} & $17\,/\,21$\\
EZSpecificity\citep{ezspecificity} & 60.18 & 57.79 & 0.6335 & 75.51 & 74.27 & 0.8686 & $16\,/\,21$\\
FusionESP\citep{fusionesp} & 60.44 & 59.35 & 0.6419 & 75.04 & 73.33 & 0.8704 & $16\,/\,21$\\
Additive ($A+C$) & 52.81 & 50.00 & 0.5055 & 53.76 & 50.00 & 0.5282 & $10\,/\,21$\\
EnzDecomp (full score) & 61.01 & 58.89 & \textbf{0.6564} & 70.63 & 68.23 & 0.8805 & $17\,/\,21$\\
$I_{\rm seq}$ (sequence branch) & \textbf{61.66} & 61.11 & 0.6481 & 75.69 & 73.89 & 0.8296 & $17\,/\,21$\\
\textbf{$I_{\rm joint}$ (Stage~3 readout)} & 61.54 & \textbf{61.63} & 0.6515 & 80.44 & \textbf{80.07} & 0.8851 & $\mathbf{19\,/\,21}$\\
\midrule
\multicolumn{8}{@{}l}{\textit{External baselines after post-hoc training-reference double centering}}\\
EnzRank + train-ref.\ centering & 48.33 & 49.02 & 0.5037 & 47.27 & 49.84 & 0.5158 & $11\,/\,21$\\
ESP + train-ref.\ centering & 57.59 & 57.26 & 0.6247 & 76.21 & 75.41 & 0.8396 & $13\,/\,21$\\
PU-EPP + train-ref.\ centering & 57.54 & 57.30 & 0.5980 & 79.02 & 79.40 & 0.8697 & $16\,/\,21$\\
RAMMESI + train-ref.\ centering & 60.18 & 59.66 & 0.6244 & 78.45 & 77.74 & 0.8581 & $12\,/\,21$\\
EZSpecificity + train-ref.\ centering & 60.30 & 59.80 & 0.6335 & 76.87 & 77.81 & 0.8686 & $\mathbf{18\,/\,21}$\\
OmniESI + train-ref.\ centering & 59.92 & 60.03 & 0.6266 & 81.35 & 81.49 & 0.8895 & $16\,/\,21$\\
ProSmith + train-ref.\ centering & 60.73 & 60.44 & \textbf{0.6473} & \textbf{83.71} & \textbf{83.99} & \textbf{0.9147} & $17\,/\,21$\\
FusionESP + train-ref.\ centering & \textbf{60.74} & \textbf{60.51} & 0.6419 & 78.64 & 78.02 & 0.8704 & $15\,/\,21$\\
\bottomrule\end{tabular}}\end{table}

\subsection{Stratified cold-isomer selectivity and stereochemical tie-collapse audit}
We called an isomer comparison cold when $s\notin S_T$ or $t\notin S_T$, with at least one substrate absent from training. This definition gave \textbf{780 high-support cold isomer cases}, comprising 326 unique pairs and 29,395 discordant enzyme comparisons (Table~\ref{tab:isomers}). The 508 constitutional-isomer cases can be distinguished by bond connectivity, and EnzRank, ESP, ProSmith and OmniESI were competitive on these. For the \textbf{237 diastereomer/epimer cases} and \textbf{35 mirror-image enantiomer cases}, the two substrates have the same 2D graph. On such comparisons the four sequence--SMILES models gave equal scores \textbf{70.5--100.0\%} of the time (raw balanced accuracy 0.00--15.60\%). With tetrahedral chirality at the candidate reaction sites and attention to 3D reference-pocket geometry, $I_{\rm joint}$ gave \textbf{0.0\% ties}. By balanced accuracy and contrast AUROC it ranked first among the native and post-hoc centered scores on \textbf{all 780 cases} (\textbf{66.29\%}, \textbf{0.7274}), on \textbf{cold diastereomers/epimers} (\textbf{58.34\%}, \textbf{0.6290}) and on \textbf{cold enantiomers} (\textbf{61.61\%}, \textbf{0.6414}).

\begin{table}[!htbp]\centering\scriptsize\setlength{\tabcolsep}{2.7pt}\renewcommand{\arraystretch}{0.96}
\caption{\textbf{Selectivity across unseen constitutional, diastereomeric/epimeric and enantiomeric substrate pairs.} All readouts and external baselines use run-averaged test score matrices across cold isomer reversal cases with $\geq 1$ unseen substrate, $\geq 20$ discordant enzymes and $\geq 20\%$ minority preference (Methods). Results scored separately by run appear as mean $\pm$ s.d.\ in Table~\ref{tab:fiverun}. In all there are 780 pair--split cases from 326 unique substrate pairs (29,395 discordant comparisons). They divide into 508 constitutional-isomer cases (185 unique pairs), 237 diastereomer/epimer cases (123 unique pairs) and 35 enantiomer cases (18 unique pairs). Tie is the exact score-tie fraction (\%), Bal is equal-case mean balanced accuracy (\%) and cAUROC is threshold-free contrast AUROC. Both mirror-image substrates were absent from training in 17 double-cold enantiomer cases (9 unique pairs and 641 comparisons). Across the 17 double-cold cases, $I_{\rm joint}$ had the top contrast AUROC ($0.5715$ versus $0.5462$ for 3D-docked EZSpecificity), with $55.09\%$ balanced accuracy compared with $56.96\%$ raw and $55.34\%$ centered for EZSpecificity. Bold marks the best value per column in each block.}\label{tab:isomers}
\resizebox{\linewidth}{!}{\begin{tabular}{@{}lrrrrrrrrrrr@{}}\toprule
& \multicolumn{3}{c}{All cold isomers (780 / 326)} & \multicolumn{2}{c}{Constitutional (508)} & \multicolumn{3}{c}{Diastereomer/epimer (237)} & \multicolumn{3}{c}{Enantiomer (35)}\\ \cmidrule(lr){2-4}\cmidrule(lr){5-6}\cmidrule(lr){7-9}\cmidrule(l){10-12}
Model / Readout & Tie & Bal (\%) & cAUROC & Bal (\%) & cAUROC & Tie & Bal (\%) & cAUROC & Tie & Bal (\%) & cAUROC\\ \midrule
\multicolumn{12}{@{}l}{\textit{Raw model readouts}}\\
EnzRank\citep{enzrank} & 34.9 & 32.59 & 0.4970 & 50.00 & 0.4951 & 100.0 & 0.07 & 0.5007 & 100.0 & 0.00 & 0.5000\\
ESP\citep{esp} & 35.1 & 44.11 & 0.6807 & 67.73 & 0.7775 & 100.0 & 0.00 & 0.5000 & 100.0 & 0.00 & 0.5000\\
OmniESI\citep{omniesi} & 24.9 & 46.95 & 0.6523 & 64.27 & 0.7281 & 71.5 & 14.46 & 0.5097 & 70.5 & 15.60 & 0.5184\\
ProSmith\citep{prosmith} & 34.0 & 46.97 & 0.7179 & \textbf{71.26} & \textbf{0.8309} & 97.5 & 1.54 & 0.5068 & 97.1 & 1.92 & 0.5083\\
EZSpecificity\citep{ezspecificity} & 0.0 & 58.67 & 0.6659 & 60.07 & 0.7162 & 0.0 & 55.55 & 0.5669 & 0.0 & 59.50 & 0.6054\\
PU-EPP\citep{puepp} & 0.0 & 59.09 & 0.6635 & 61.67 & 0.7214 & 0.0 & 54.40 & 0.5575 & 0.0 & 53.45 & 0.5410\\
RAMMESI\citep{rammesi} & 0.0 & 59.39 & 0.6607 & 61.94 & 0.7106 & 0.0 & 54.61 & 0.5698 & 0.0 & 54.89 & 0.5521\\
FusionESP\citep{fusionesp} & 0.0 & 63.10 & 0.7200 & 66.38 & 0.7780 & 0.0 & 56.74 & 0.6088 & 0.0 & 58.56 & 0.6298\\
Additive ($A+C$) & 0.0 & 50.00 & 0.5063 & 50.00 & 0.5075 & 0.0 & 50.00 & 0.5045 & 0.0 & 50.00 & 0.5018\\
EnzDecomp (full score) & 0.0 & 63.59 & 0.7242 & 66.85 & 0.7747 & 0.0 & 57.25 & 0.6286 & 0.0 & 59.17 & 0.6383\\
\textbf{$I_{\rm joint}$ (Stage~3 readout)} & 0.0 & \textbf{66.29} & \textbf{0.7274} & 70.32 & 0.7792 & 0.0 & \textbf{58.34} & \textbf{0.6290} & 0.0 & \textbf{61.61} & \textbf{0.6414}\\
\midrule
\multicolumn{12}{@{}l}{\textit{External baselines after post-hoc training-reference double centering}}\\
EnzRank + train-ref.\ centering & 33.5 & 33.13 & 0.4973 & 49.26 & 0.4956 & 95.8 & 3.02 & 0.5007 & 94.3 & 2.86 & 0.5000\\
ESP + train-ref.\ centering & 35.0 & 45.93 & 0.6808 & 70.52 & 0.7776 & 100.0 & 0.00 & 0.5000 & 100.0 & 0.00 & 0.5000\\
OmniESI + train-ref.\ centering & 0.3 & 61.28 & 0.6488 & 67.08 & 0.7281 & 0.8 & 50.43 & 0.5002 & 0.0 & 50.55 & 0.5031\\
EZSpecificity + train-ref.\ centering & 0.0 & 61.63 & 0.6659 & 64.99 & 0.7162 & 0.0 & 54.96 & 0.5669 & 0.0 & 57.97 & 0.6054\\
RAMMESI + train-ref.\ centering & 0.0 & 61.88 & 0.6607 & 65.80 & 0.7106 & 0.0 & 54.94 & 0.5698 & 0.0 & 52.03 & 0.5521\\
PU-EPP + train-ref.\ centering & 0.0 & 61.93 & 0.6635 & 66.58 & 0.7214 & 0.0 & 53.29 & 0.5575 & 0.0 & 53.05 & 0.5410\\
ProSmith + train-ref.\ centering & 5.5 & 64.36 & 0.7184 & \textbf{76.03} & \textbf{0.8309} & 15.6 & 42.66 & 0.5086 & 17.1 & 41.90 & 0.5075\\
FusionESP + train-ref.\ centering & 0.0 & \textbf{65.21} & \textbf{0.7200} & 69.19 & 0.7780 & 0.0 & \textbf{57.53} & \textbf{0.6088} & 0.0 & \textbf{59.38} & \textbf{0.6298}\\
\bottomrule\end{tabular}}\end{table}

\clearpage
{\setlength{\intextsep}{5pt plus 2pt minus 2pt}\setlength{\textfloatsep}{6pt plus 2pt minus 2pt}
\section{Computational footprint, internal ablations and single-run stability}\label{sec:supp_ablations}

\subsection{Downstream GPU training time and memory profile}
Once a substrate has been encoded, its reaction-site descriptors can be used for all the enzymes in the screen. An enzyme's reference active-site pocket can likewise be used for all the substrates. This avoids $N_E\times N_S$ separate 3D docking calculations. On the 40-split esterase screen, five fits and their validation calibration after feature caching cost \textbf{10.44~min} per split (\textbf{2.09~min} per run) and \textbf{3.13~GB} of peak GPU memory on NVIDIA A800-SXM4-80GB hosts (Table~\ref{tab:runtime} and Fig.~\ref{fig:mechanism}d). On one split of downstream training, OmniESI used 22.40~min, so EnzDecomp ran \textbf{2.1$\times$ faster}. EZSpecificity, which trains on 3D docked complexes, used 263.45~min and 9.59~GB. Relative to that model, EnzDecomp was \textbf{25$\times$ faster} and used \textbf{3.1$\times$ less memory}. ProSmith used 459.60~min and 62.85~GB. EnzDecomp was \textbf{44$\times$ faster} and used \textbf{20$\times$ less memory}. On frozen features, ESP's gradient-boosted trees and EnzRank's CNNs ran faster at 3.90--4.95~min per split, but their screening and stereoisomer accuracies were substantially lower than EnzDecomp's (Tables~\ref{tab:overall}--\ref{tab:isomers}).

\begin{table}[!htbp]\centering\scriptsize\setlength{\tabcolsep}{3.2pt}
\caption{\textbf{Elapsed training time and peak device memory on the esterase panel.} Downstream training jobs were timed on NVIDIA A800-SXM4-80GB hosts. Timings exclude cached-feature preparation but cover the full downstream job, including GPU-external overhead. In Splits (runs), the first value is the number of splits with A800 timing records and the value in parentheses is the number of individual training runs recorded. Min/run is the mean duration of one run. Min/split ($=5\times\text{Min/run}$, Methods) is reported to enable comparison on a five-run split. $^\ast$For EnzDecomp, all five repeats across all 40 splits were executed consecutively within the same job, with a mean job duration of 10.44~min, equivalent to 2.09~min per run.}\label{tab:runtime}
\resizebox{\linewidth}{!}{\begin{tabular}{@{}lcrrrl@{}}\toprule
Model & Splits (runs) & Min/split & Min/run & Peak VRAM (GB) & Timed downstream work\\ \midrule
\textbf{EnzDecomp (ours)} & 40 (40$^\ast$) & 10.44 & 2.09 & 3.13 & Three-stage fit, sequence branch and validation calibration\\
EZSpecificity\citep{ezspecificity} & 40 (178) & 263.45 & 52.69 & 9.59 & 3D complex graph network training and validation (optimized tier)\\
ProSmith\citep{prosmith} & 40 (189) & 459.60 & 91.92 & 62.85 & Multimodal Transformer fine-tuning and gradient-boosted trees\\
RAMMESI\citep{rammesi} & 40 (180) & 178.10 & 35.62 & 54.56 & Retrieval-augmented downstream training (optimized; 186.45 min/run native)\\
PU-EPP\citep{puepp} & 40 (183) & 114.30 & 22.86 & 13.54 & Positive--unlabeled downstream training (optimized; 84.08 min/run native)\\
FusionESP\citep{fusionesp} & 40 (190) & 34.00 & 6.80 & 0.36 & Feature-fusion network training and validation\\
OmniESI\citep{omniesi} & 40 (177) & 22.40 & 4.48 & 2.80 & Conditional network training and validation (optimized; 4.73 min/run native)\\
EnzRank\citep{enzrank} & 11 (11) & 4.95 & 0.99 & 0.46 & CNN and fingerprint network training (1.48 min/run over 181 optimized runs)\\
ESP\citep{esp} & 7 (7) & 3.90 & 0.78 & $<0.10$ & Gradient-boosted tree fitting (0.08 min/run over 185 optimized runs)\\
\bottomrule\end{tabular}}\vspace{-0.5em}\end{table}

\subsection{Component and training ablations}
Across the 176 splits in Table~\ref{tab:internal}, scoring reactions by enzyme breadth $A(e)$ and shared substrate reactivity $C(s)$ alone ($A+C$) gave \textbf{0.4683 Row AP}, compared with 0.3451 for $I_{\rm joint}$ alone. We next let the additive fit make an enzyme-specific correction by adding a scalar times the centered Stage~3 residual, giving $A+C_r+\gamma J$. With no sequence-only prior and no post-hoc calibration, this three-stage model gave \textbf{0.4781 Row AP}, higher than each external score. Alignment on the validation data according to support brought the full score to \textbf{0.4830 Row AP} (Tables~\ref{tab:internal} and~\ref{tab:internal_reversal}). We fit enzyme breadth $A(e)$, substrate reactivity $C(s)$ and the pocket interaction $I(e,s)$ separately during three-stage training. For comparison, we centered the scores of an end-to-end network after training (Fig.~\ref{fig:mechanism}a,b). We trained the same network from scratch without freezing the marginal terms and tested it on the 105 held-out esterase substrate pairs (Table~\ref{tab:internal_reversal}). Balanced reversal accuracy was \textbf{59.63\%} before double centering on the training reference panel and \textbf{68.01\%} afterwards. We froze $C_r(s)$ and then $A(e)$. The interaction layers fit only the pocket residual. With active-site sequence shells alone, $I_{\rm seq}$ reached \textbf{73.89\%} balanced reversal accuracy. With attention from reaction sites to 3D reference-pocket blocks, $I_{\rm joint}$ reached \textbf{80.07\%} and made the correct prediction in \textbf{19/21} comparisons of isomers \#4 and \#14. The staged uncentered residual ($A+C_r+\gamma D$) gave 50.77\% balanced accuracy before post-hoc centering and 61.19\% afterwards. The accuracy of $I_{\rm joint}$ depended on both sequential freezing and centering during training.

\begin{table}[!htbp]\centering\footnotesize
\caption{\textbf{Library ranking with components omitted or the training procedure changed.} The values for each model are averages over 176 test splits, with equal weight per split. For Column AP and AUROC there are 175 defined values. Metrics were calculated from run-averaged test scores (Methods). The largest mean in each column is in bold.}\label{tab:internal}
\begin{tabular}{@{}lrrrrr@{}}\toprule
Readout / Ablation & Splits & Row AP & Column AP & Pooled AP & AUROC\\ \midrule
Sequence interaction ($I_{\rm seq}$) & 176 & 0.3589 & 0.4277 & 0.3281 & 0.5975\\
Joint interaction ($I_{\rm joint}$) & 176 & 0.3451 & 0.4185 & 0.3150 & 0.5793\\
Unstaged end-to-end + post hoc reference centering & 176 & 0.3380 & 0.3764 & 0.2671 & 0.5380\\
Staged uncentered residual + post hoc reference centering & 176 & 0.3341 & 0.3878 & 0.2796 & 0.5491\\
Additive main effects ($A+C$) & 176 & 0.4683 & 0.5261 & 0.4875 & 0.7731\\
Staged uncentered residual ($A+C_r+\gamma D$) & 176 & 0.4768 & 0.5332 & 0.5007 & 0.7772\\
Staged base ($A+C_r+\gamma J$) & 176 & 0.4781 & 0.5336 & 0.4999 & 0.7765\\
Unstaged joint end-to-end (full score) & 176 & 0.4785 & 0.5343 & \textbf{0.5218} & \textbf{0.7847}\\
\textbf{EnzDecomp (leave-one-source-out rule selection)} & 176 & 0.4829 & 0.5449 & 0.5188 & 0.7782\\
\textbf{EnzDecomp (full score)} & 176 & \textbf{0.4830} & \textbf{0.5453} & 0.5202 & 0.7782\\
\bottomrule\end{tabular}\vspace{-0.4em}\end{table}
\begin{table}[!htbp]\centering\scriptsize\setlength{\tabcolsep}{3.2pt}
\caption{\textbf{Substrate preference across ablations of the same backbone.} The first block reports all 4,355 reversal cases across panels (3,127 unique pairs) using run-averaged test scores (Methods). The second block reports the 105 esterase substrate pairs from substrate holdouts (3,472 comparisons) and the unseen constitutional isomer pair \#4 vs.\ \#14. On the 105 esterase substrate pairs, reference centering after training gave the unstaged joint score 67.60\% direction accuracy and 68.01\% balanced accuracy ($13\,/\,21$ on \#4 vs.\ \#14). For the staged uncentered residual, centering after training gave 62.34\% direction accuracy and 61.19\% balanced accuracy ($16\,/\,21$). Both remain well below double centering during training in $I_{\rm joint}$ (80.44\% and 80.07\%).}\label{tab:internal_reversal}
\resizebox{\linewidth}{!}{\begin{tabular}{@{}lrrrrrrr@{}}\toprule
& \multicolumn{3}{c}{All cross-panel cases (4,355)} & \multicolumn{4}{c}{Esterase substrate holdout (105 pairs)}\\ \cmidrule(lr){2-4}\cmidrule(l){5-8}
Readout / Ablation & Dir (\%) & Bal (\%) & cAUROC & Dir (\%) & Bal (\%) & cAUROC & \#4 vs.\ \#14\\ \midrule
Additive baseline ($A+C$) & 52.81 & 50.00 & 0.5055 & 53.76 & 50.00 & 0.5282 & $10\,/\,21$\\
Staged uncentered residual ($A+C_r+\gamma D$) & 58.93 & 56.40 & 0.6308 & 54.62 & 50.77 & 0.5834 & $10\,/\,21$\\
Staged base ($A+C_r+\gamma J$) & 59.66 & 57.60 & 0.6460 & 59.52 & 55.94 & 0.7982 & $16\,/\,21$\\
Unstaged joint end-to-end (raw) & 57.91 & 55.51 & 0.6072 & 64.79 & 59.63 & 0.7504 & $16\,/\,21$\\
\textbf{EnzDecomp (LOSO rule selection)} & 61.00 & 58.88 & \textbf{0.6564} & 70.63 & 68.23 & 0.8805 & $17\,/\,21$\\
\textbf{EnzDecomp (full library-ranking score)} & 61.01 & 58.89 & \textbf{0.6564} & 70.63 & 68.23 & 0.8805 & $17\,/\,21$\\
Sequence interaction readout ($I_{\rm seq}$) & \textbf{61.66} & 61.11 & 0.6481 & 75.69 & 73.89 & 0.8296 & $17\,/\,21$\\
\textbf{EnzDecomp interaction readout ($I_{\rm joint}$)} & 61.54 & \textbf{61.63} & 0.6515 & \textbf{80.44} & \textbf{80.07} & \textbf{0.8851} & $\mathbf{19\,/\,21}$\\
\bottomrule\end{tabular}}\vspace{-0.9em}\end{table}

\vspace{-0.45em}
\subsection{Variation across individual runs}
For the main text and figures we averaged the test scores over runs for each model before calculating metrics (Methods). Here each external model and each EnzDecomp readout, including $I_{\rm joint}$, is scored separately, one run at a time, across the five repeated runs for all ten headline benchmarks (Table~\ref{tab:fiverun}). Table entries report the mean $\pm$ s.d.\ across runs, averaged with equal weight over cases, with the highest mean in each column shown in bold. Rankings from every individual run agree with the multi-run averages reported in Tables~\ref{tab:overall}--\ref{tab:isomers}. Single-run $\hat f_{\rm EnzDecomp}$ ranks \textbf{first} for Row AP across all 176 splits ($\mathbf{.4694\pm.0048}$), 152 OOD splits ($\mathbf{.4604\pm.0057}$), and 74 double-holdout splits ($\mathbf{.4182\pm.0092}$), and leads Pooled AP ($\mathbf{.2922\pm.0073}$) and AUROC ($\mathbf{.6841\pm.0084}$) across 40 curated HAD splits. Single-run $I_{\rm joint}$ ranks \textbf{first} across all 4,355 reversal cases ($\mathbf{59.55\pm0.57\%}$), 780 cold isomers ($\mathbf{63.96\pm0.43\%}$), 237 cold epimers ($\mathbf{57.28\pm0.52\%}$), and 35 cold enantiomers ($\mathbf{57.29\pm1.51\%}$).

\begin{table}[!htbp]\centering\scriptsize\setlength{\tabcolsep}{2.2pt}\renewcommand{\arraystretch}{0.96}
\caption{\textbf{Variation across individual runs for EnzDecomp and the external predictors.} For the main text, figures and Tables~\ref{tab:overall}--\ref{tab:isomers} we averaged the test scores over runs for each model before calculating metrics (Methods). Here each external model and each EnzDecomp readout, including $I_{\rm joint}$, is scored separately, one run at a time. Values are means across runs ($\pm$ s.d.), with the highest mean in each column in bold and equal weight for each case. Activity screening columns report Row AP and Pooled AP across all 176 splits, Row AP across 152 out-of-distribution (OOD) and 74 double-holdout splits, and Pooled AP and AUROC across 40 curated HAD phosphatase splits. Selectivity reversal columns report mean balanced accuracy (\%) before and after double-centering on each split's training reference panel for all 4,355 cross-panel reversals and 780 cold-isomer cases, with separate results for 237 cold diastereomer/epimer cases and 35 cold enantiomer cases.}\label{tab:fiverun}
\resizebox{\linewidth}{!}{\begin{tabular}{@{}lrrrrrrrrrr@{}}\toprule
& \multicolumn{6}{c}{Activity screening (single-run mean $\pm$ s.d.)} & \multicolumn{4}{c}{Reversal balanced accuracy (\%, single-run mean $\pm$ s.d.)}\\ \cmidrule(lr){2-7}\cmidrule(l){8-11}
Model / Readout & 176 Row & 176 Pool & 152 OOD & 74 Dual & HAD Pool & HAD AUC & All (4,355) & Iso.\ (780) & Epim.\ (237) & Enant.\ (35)\\ \midrule
\multicolumn{11}{@{}l}{\textit{Single-run raw baselines and EnzDecomp readouts}}\\
FusionESP\citep{fusionesp} & $.4173\pm.0031$ & $.4346\pm.0051$ & $.4042\pm.0033$ & $.3601\pm.0076$ & $.2153\pm.0070$ & $.6165\pm.0100$ & $57.64\pm0.60$ & $61.81\pm1.54$ & $55.82\pm2.02$ & $56.96\pm4.26$\\
PU-EPP\citep{puepp} & $.4345\pm.0071$ & $.4350\pm.0038$ & $.4228\pm.0084$ & $.3811\pm.0117$ & $.2166\pm.0038$ & $.5977\pm.0063$ & $54.00\pm0.85$ & $56.42\pm1.14$ & $52.06\pm1.51$ & $52.09\pm1.53$\\
EnzRank\citep{enzrank} & $.4441\pm.0092$ & $.4505\pm.0037$ & $.4380\pm.0102$ & $.4111\pm.0119$ & $.2250\pm.0087$ & $.6133\pm.0165$ & $48.45\pm0.00$ & $32.58\pm0.01$ & $0.05\pm0.03$ & $0.00\pm0.00$\\
ESP\citep{esp} & $.4381\pm.0011$ & $.4826\pm.0010$ & $.4257\pm.0012$ & $.3733\pm.0029$ & $.2679\pm.0006$ & $.6589\pm.0005$ & $55.04\pm0.02$ & $44.01\pm0.03$ & $0.00\pm0.00$ & $0.00\pm0.00$\\
RAMMESI\citep{rammesi} & $.4573\pm.0038$ & $.5030\pm.0042$ & $.4453\pm.0043$ & $.3954\pm.0050$ & $.2617\pm.0011$ & $.6574\pm.0029$ & $55.73\pm0.67$ & $57.62\pm0.54$ & $53.40\pm0.97$ & $52.84\pm1.74$\\
EZSpecificity\citep{ezspecificity} & $.4633\pm.0022$ & $.5071\pm.0021$ & $.4531\pm.0025$ & $.4132\pm.0042$ & $.2543\pm.0059$ & $.6468\pm.0060$ & $57.32\pm0.95$ & $56.89\pm0.81$ & $52.50\pm1.19$ & $54.67\pm2.70$\\
OmniESI\citep{omniesi} & $.4612\pm.0062$ & $.5205\pm.0046$ & $.4502\pm.0066$ & $.4024\pm.0106$ & $.2620\pm.0065$ & $.6654\pm.0056$ & $55.45\pm1.15$ & $44.93\pm0.40$ & $12.20\pm0.55$ & $13.50\pm0.76$\\
ProSmith\citep{prosmith} & $.4649\pm.0026$ & $\mathbf{.5227}\pm.0032$ & $.4537\pm.0031$ & $.4104\pm.0043$ & $.2449\pm.0026$ & $.6486\pm.0026$ & $56.77\pm0.17$ & $46.30\pm0.28$ & $0.74\pm0.20$ & $0.99\pm0.28$\\
\textbf{EnzDecomp (full, single-run)} & $\mathbf{.4694}\pm.0048$ & $.5057\pm.0060$ & $\mathbf{.4604}\pm.0057$ & $\mathbf{.4182}\pm.0092$ & $\mathbf{.2922}\pm.0073$ & $\mathbf{.6841}\pm.0084$ & $57.58\pm0.63$ & $61.09\pm0.55$ & $54.66\pm0.86$ & $55.58\pm1.45$\\
\textbf{$I_{\rm joint}$ (single-run)} & $.3364\pm.0050$ & $.2920\pm.0065$ & $.3273\pm.0055$ & $.3077\pm.0068$ & $.1481\pm.0040$ & $.4954\pm.0064$ & $\mathbf{59.55}\pm0.57$ & $\mathbf{63.96}\pm0.43$ & $\mathbf{57.28}\pm0.52$ & $\mathbf{57.29}\pm1.51$\\
\midrule
\multicolumn{11}{@{}l}{\textit{Single-run external baselines + post-hoc training-reference double centering (selectivity readouts)}}\\
EnzRank + train-ref.\ centering & --- & --- & --- & --- & --- & --- & $48.66\pm0.27$ & $33.39\pm0.30$ & $2.94\pm0.09$ & $2.57\pm0.64$\\
ESP + train-ref.\ centering & --- & --- & --- & --- & --- & --- & $57.26\pm0.08$ & $45.89\pm0.02$ & $0.00\pm0.00$ & $0.00\pm0.00$\\
PU-EPP + train-ref.\ centering & --- & --- & --- & --- & --- & --- & $55.57\pm0.71$ & $58.24\pm1.05$ & $51.90\pm1.61$ & $51.79\pm1.44$\\
RAMMESI + train-ref.\ centering & --- & --- & --- & --- & --- & --- & $57.80\pm0.79$ & $59.14\pm0.92$ & $53.44\pm1.21$ & $52.04\pm2.83$\\
OmniESI + train-ref.\ centering & --- & --- & --- & --- & --- & --- & $58.46\pm0.70$ & $59.30\pm0.35$ & $49.77\pm0.63$ & $50.88\pm0.27$\\
EZSpecificity + train-ref.\ centering & --- & --- & --- & --- & --- & --- & $58.92\pm1.13$ & $59.22\pm0.51$ & $52.41\pm1.08$ & $54.89\pm2.44$\\
ProSmith + train-ref.\ centering & --- & --- & --- & --- & --- & --- & $59.06\pm0.49$ & $57.86\pm1.75$ & $25.49\pm4.19$ & $28.71\pm6.48$\\
FusionESP + train-ref.\ centering & --- & --- & --- & --- & --- & --- & $58.82\pm0.66$ & $63.63\pm1.54$ & $56.29\pm2.25$ & $56.98\pm5.24$\\
\bottomrule\end{tabular}}\vspace{-0.7em}\end{table}

\section{Reproducibility statement}\label{sec:repro}
We saved all 176 train--validation--test split manifests, standardized substrate structures and InChIKey classifications for all 870 unique isomer pairs. We cached ESM-C residue shells, ATOMICA 3D reference-pocket blocks and 160-channel reaction-site chirality and GFN2-xTB/AIMNet2 descriptors. Trained checkpoints accompany 64-bit floating-point stage decompositions that agree with the full readout to numerical precision ($\max|\Delta|=8.9\times 10^{-16}$, Fig.~\ref{fig:selectivity}c). Single-run and run-averaged test score matrices cover EnzDecomp and eight external baselines. Scripts reproduce Tables~\ref{tab:panels}--\ref{tab:fiverun}, Figs.~\ref{fig:architecture}--\ref{fig:mechanism} and Fig.~\ref{fig:s1}. All these materials will be made publicly available upon publication and can be provided to editors and reviewers upon request during peer review.\par}

\end{document}